\documentclass[final,1p,times]{elsarticle}
\usepackage{helvet}
\usepackage{courier}

\usepackage{times}
\usepackage{latexsym}
\usepackage{graphicx}
\usepackage{color}
\usepackage{multirow}
\usepackage{wrapfig}
\usepackage[usenames,dvipsnames]{xcolor}
\usepackage[ruled,vlined,linesnumbered]{algorithm2e}
\usepackage{float}
\usepackage{tabularx}
\usepackage{multicol}

\usepackage{natbib}
\usepackage{geometry}
\usepackage{pifont}

\newtheorem{definition}{Definition}

\newcommand{\comment}[1]{}

\journal{axXiv}

\begin{document}

\begin{frontmatter}


\title{A Summary of Adaptation of Techniques from Search-based Optimal Multi-Agent Path Finding Solvers to Compilation-based Approach }

\author{Pavel Surynek}
\ead{pavel.surynek@fit.cvut.cz}

\address{Faculty of Information Technology, Czech Technical University in Prague,\\Th\'{a}kurova 9, 160 00 Praha 6, Czech Republic}




\begin{abstract}

In the \emph{multi-agent path finding} problem (MAPF) we are given a set of
agents each with respective start and goal positions. The task is to find paths
for all agents while avoiding collisions aiming to minimize an objective
function. Two such common objective functions is the {\em sum-of-costs} and the {\em
makespan}. Many optimal solvers were introduced in the past decade - two prominent categories
of solvers can be disntinguished: {\em search-based} solvers and {\em compilation-based} solvers.

Search-based solvers were developed and tested for the sum-of-costs objective while
the most prominent compilation-based solvers that are built around Boolean satisfiability (SAT) were designed for the makespan objective. Very little was known on the performance and relevance of the compilation-based approach on the sum-of-costs objective.

In this paper we show how to close the gap between these cost functions in the compilation-based approach. Moreover we study
applicability of various techniuqes developed for search-based solvers in the compilation-based approach.

A part of this paper introduces a SAT-solver that is directly
aimed to solve the sum-of-costs objective function. Using both a lower bound on
the sum-of-costs and an upper bound on the makespan, we are able to have a
reasonable number of variables in our SAT encoding. We then further improve the
encoding by borrowing ideas from \textsc{Icts}, a search-based solver.

Experimental evaluation on several domains show that there are many scenarios where our new
SAT-based methods outperforms the best variants of previous sum-of-costs search
solvers - the \textsc{Icts}, \textsc{Cbs} algorithms, and \textsc{Icbs} algorithms.
\end{abstract}

\begin{keyword}
Multi-agent path finding (MAPF), sum-of-costs, makespan, Boolean satisfiability (SAT), optimality, suboptimality, propositional encoding, cardinality constraint
\end{keyword}

\end{frontmatter}

\section{Introduction and Background}
\noindent
The {\em multi-agent path finding} (MAPF) problem consists  a graph, $G=(V,E)$
and a set $A=\{a_1, a_2,\dots a_k\}$ of $k$ agents. Time is discretized into
time steps. The arrangement of agents at time-step $t$ is denoted as
$\alpha_t$. Each agent $a_i$ has a start position $\alpha_0(a_i) \in V$ and a
goal position $\alpha_+(a_i) \in V$.  At each time step an agent can either
{\em move} to an adjacent empty location\footnote{Some variants of MAPF relax
the empty location requirement by allowing a chain of neighboring agents to
move, given that the
head of the chain enters an empty locations.
Most MAPF algorithms are robust (or at least easily modified) across these
variants.} or {\em wait} in its current location. The task is to find a
sequence of move/wait actions for each agent $a_i$, moving it from
$\alpha_0(a_i)$ to $\alpha_+(a_i)$ such that agents do not {\em conflict},
i.e., do not occupy the same location at the same time. Formally, an MAPF
instance is a tuple $\Sigma=(G=(V,E),A,\alpha_0,\alpha_+)$. A \textit{solution}
for $\Sigma$ is a sequence of arrangements
$\mathcal{S}(\Sigma)=[\alpha_0,\alpha_1,...,\alpha_{\mu}]$ such that
$\alpha_{\mu}=\alpha_+$ where $\alpha_{t+1}$ results from valid movements from
$\alpha_{t}$ for $t=1,2,...,\mu-1$. An example of MAPF and its solution are
shown in Figure \ref{figure-MAPF}.

MAPF has practical applications in video games, traffic control, robotics etc.
(see \cite{CBSJUR} for a survey). The scope of this paper is limited to the
setting of {\em fully cooperative} agents that are centrally controlled. MAPF
is usually solved aiming to minimize one of the two commonly-used global
cumulative cost functions: \vspace{0.25cm}

\begin{figure}[t]
\centering
\includegraphics[trim={2.5cm 22.6cm 3cm 2.5cm},clip,width=1.0\textwidth]{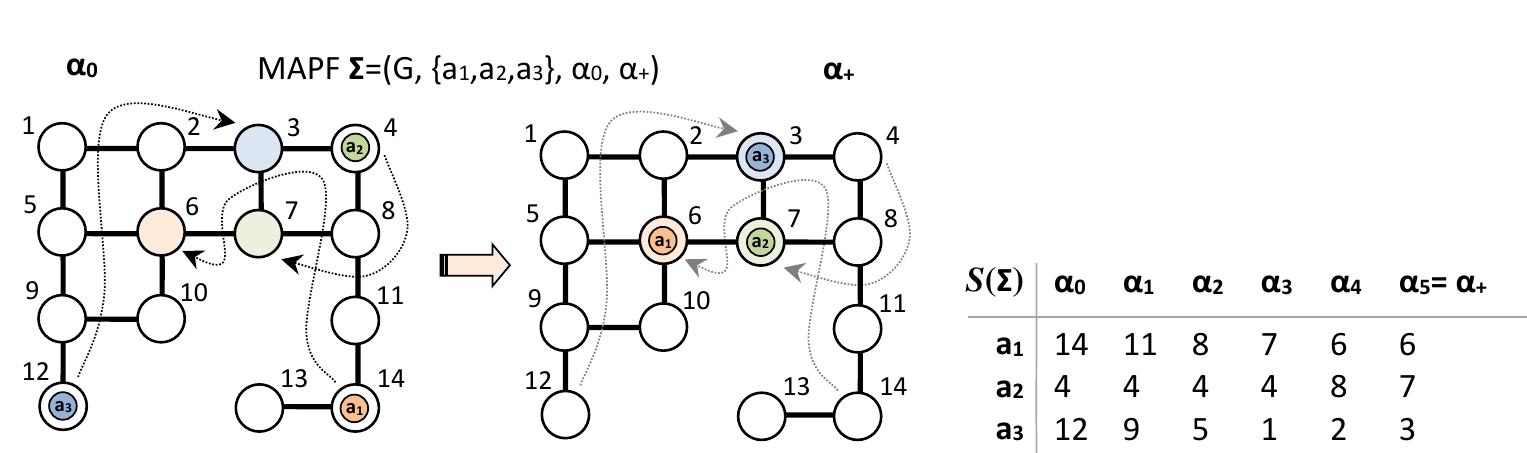}
\caption{Example of MAPF for agents $a_1$, $a_2$, and $a_3$ over a
$4$-connected grid (left) and its solution (right).} \label{figure-MAPF}
\end{figure}

\noindent{\bf (1)  sum-of-costs} (denoted $\xi$) is the summation, over all
agents, of the number of time steps required to reach the goal
location~\cite{dresner2008aMultiagent,standley2010finding,DBLP:journals/ai/SharonSGF13,CBSJUR}.
Formally, $\xi = \sum_{i=1}^k{\xi(a_i)}$, where $\xi(a_i)$ is an
\textit{individual path cost} of agent $a_i$.
\vspace{0.25cm}

\noindent{\bf (2) makespan:} (denoted $\mu$) is the total time until the last
agent reaches its destination (i.e., the maximum of the individual
costs)~\cite{DBLP:conf/aaai/Surynek10,DBLP:conf/ictai/Surynek14,DBLP:conf/ijcai/Surynek15}.
\vspace{0.25cm}

In the indivudual path cost, each action of an agent (move action or wait action) is assumed to have a unit cost. It is important to note that in any solution $\mathcal{S}(\Sigma)$ it holds that $\mu \leq \xi \leq {m\cdot\mu}$ Thus the optimal \textit{makespan} is
usually smaller than the optimal \textit{sum-of-costs}.

\begin{figure}[t]
\centering
\includegraphics[trim={3cm 18.9cm 2.5cm 2cm},clip,width=1.0\textwidth]{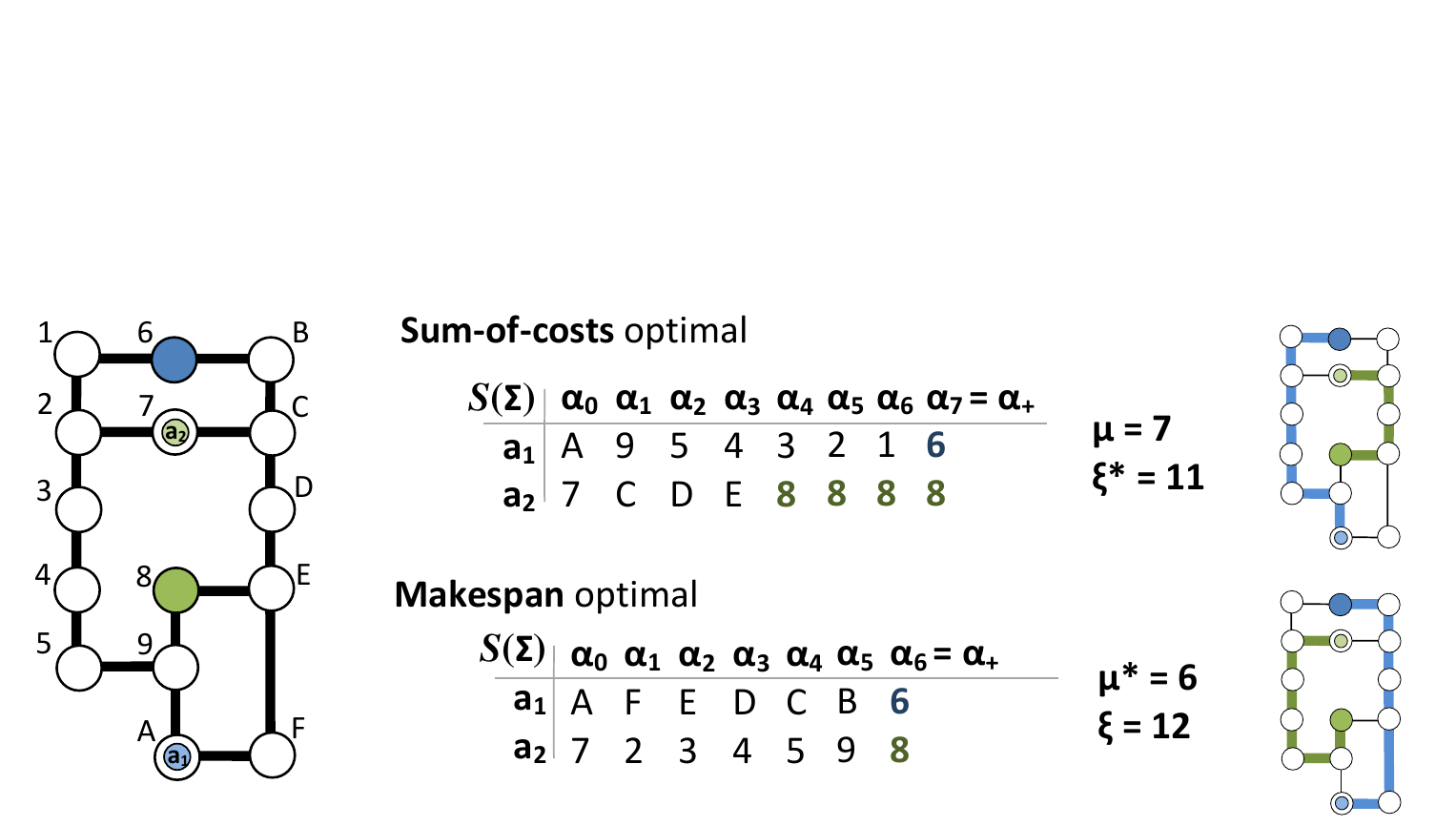}
\caption{An instance of MAPF where {\em makespan} and {\em
sum-of-costs} optimal solutions differ - that is, any makespan optimal solution
is strictly sum-of-costs suboptimal and any sum-of-costs optimal solution is strictly makespan suboptimal.}
\label{figure-MAKEvsCOST}
\end{figure}

Intuitivelly, sum-of-costs can be regarded as total energy consumption of all agents such that at each time step spent before reaching the goal the agent consumes one unit of energy. In this respect, it is not surprising that optimization of one of these two objectives goes against the other - total time can be saved at the cost of increased energy consumption and vice versa. An example of MAPF instance where any makespan optimal solution has sum-of-costs that is greater than the optimum and any sum-of-costs optimal solution has makespan that is greater than the optimal makespan is shown in Figure \ref{figure-MAKEvsCOST}.

Finding optimal solutions for both variants with any standard style of agents' movement is
NP-hard even on planar graphs~\cite{DBLP:conf/aaai/RatnerW86,DBLP:journals/jsc/RatnerW90,DBLP:conf/aaai/Surynek10,DBLP:journals/fuin/Surynek15,DBLP:conf/aaai/YuL13,DBLP:journals/ral/Yu16}. Therefore, many suboptimal solvers were developed and are usually used when $m$ is large or when the graph is large~\cite{CohenUK15,KhorshidHS11,roger2012non,DBLP:conf/icra/Ryan10,DBLP:conf/aiide/Silver05,DBLP:journals/jair/WangB11}. In contrast to difficulty of finding optimal solutions, finding any feasible solution or detecting unsolvability of a given instance can be done polynomial time \cite{DBLP:journals/jair/WildeMW14,DBLP:conf/focs/KornhauserMS84,DBLP:conf/aaai/LunaB11,DBLP:conf/icra/Surynek09,DBLP:journals/ci/Surynek14}.

\subsection{Optimal MAPF Solvers}

Many optimal solvers were introduced in the past decade but they all focus on
one of these cost functions:

\begin{itemize}

\item {\bf (1) optimal sum-of-costs solvers.} Most of them are based on
search. Some of these search-based solvers are variants of the A* algorithm on
a global {\em search space} in which all different ways to place $m$ agents into $V$
vertices, one agent per vertex, are considered~\cite{standley2010finding,DBLP:journals/ai/WagnerC15}.
Other employ novel search trees~\cite{DBLP:conf/ijcai/BoyarskiFSSTBS15,CBSJUR,DBLP:journals/ai/SharonSGF13}.
Search-based solvers feature various search space compilation techniques like {\em independence detection} (ID) \cite{standley2010finding}
or {\em multi-value decision diagrams} (MDDs) \cite{DBLP:journals/ai/SharonSGF13}.

\item {\bf (2) optimal makespan solvers}. Many optimal solvers were developed
for the makespan variant. Most of them are {\em compilation-based solvers} which
reduce MAPF to known problems such as Constraint Satisfaction (CSP)~\cite{DBLP:conf/icra/Ryan10},
Boolean Satisfiability (SAT)\cite{surynek2012towards}, Inductive Logic
Programming (ILP) \cite{yu2013planning} and Answer Set
Programming (ASP) \cite{erdem2013general}. These works mostly prove a
polynomial-time reduction from MAPF to these problems. Existing reductions are
usually designed for the {\em makespan} variant of MAPF; they are not
applicable for the sum-of-costs variant.

\end{itemize}

\subsection{Current Shortcomings and Contribution}

A major weaknesses across all these works is that each of these algorithms was
introduced and applied for one of these objective functions only. Furthermore,
the connection/comparison between different algorithms was usually done only
within a given class of algorithms and cost objective but not across these two
classes. Finally, experiments were always performed on one objective-function
and very little is known on the performance and relevance of any given
algorithm (developed for one cost function) on the other objective function.

This paper aims to close the gap. First, we discuss how to migrate
algorithms across the different objective functions. Most of the search-based
algorithms developed for the sum-of-cost objective function can be modified to
the makespan variant with some technical adaptations such as modifying the cost
function and the way the state-space is represented. Some initial directions
are given by~\cite{CBSJUR} and we give a complete picture here.

By contrast, the compilation-based algorithms that were developed for the
makespan objective function are not trivially modified to the sum-of-costs
variant and sometimes a completely new encoding is needed.

A major algorithmic contribution of this paper is that we develop the compilation-based solver
for the sum-of-costs variant to SAT. Our SAT-based solver is based on establishing relations between
the maximum makespan under the given sum-of-costs which enables to build SAT encodings that represent
all feasible solutions for the given sum-of-costs. Bounds on the sum-of-costs in the SAT encoding are established by
{\em cardinality constraints}~\cite{DBLP:conf/cp/BailleuxB03,DBLP:conf/cp/SilvaL07}. We show how to use known
lower bounds on the sum-of-costs to reduce the number of variables that encode these cardinality constraints so as
to be practical for current SAT solvers.

We then present how to migrate various techniques used in search-based approach to our new SAT-based solver.
First, we adapt ideas from the \textsc{Icts} algorithm~\cite{DBLP:journals/ai/SharonSGF13} that uses {\em multi-value
decision diagrams} (MDDs)~\cite{DBLP:conf/iccad/SrinivasanKMB90} to further
reduce the size of SAT encodings. Next, we show how to integrate a modification of {\em independece detection} \cite{standley2010finding} technique into the SAT-based solver. Finally, we demonstrate flexibility of out SAT-based solver by modifying it into a bounded sum-of-costs {\em suboptimal} solver - a modification applicable in search-based approach to trade-off quality of solutions and runtime \cite{DBLP:conf/ecai/BarerSSF14}.

Successful migration of techniques demonstrates the potential of combining ideas from
both classes of approaches - search-based and compilation-based. Experimental results
show that our SAT solver with various enhancements outperforms the best existing search-based
solvers for the sum-of-costs variant on a number scenarios.

Hence as a results of our unification provided in the beginning of this paper
we have an arsenal of algorithms which can be applied for both objective
functions. We conclude this paper by providing experimental results comparing
the hardness of solving MAPF with SAT-based and search-based solvers under the
makespan and the sum-of-costs objectives in a number of domains.

\section{Related Work}

We summarize existing algorithmic approaches to MAPF in this section. We categorize algorithms into two streams according to the objective function they use. For optimization of {\em sum-of-costs} great variety of algorithms has been proposed. On the other hand, previous {\em makespan} optimal algorithms are limited to compilation-based approach where the target formalism is represented by Boolean satisfiability. Many sum-of-costs optimal algorithms can be directly modified for the makespan variant. The opposite migration from the makespan optimal case to sum-of-costs optimality in compilation-based algorithms is however not straightforward. 

\subsection{Previous Sum-of-Costs Optimal Algorithms and Techniques}

{\bf A*-based Algorithms}. A* is a general-purpose algorithm that is well suited to solve MAPF. A common
straightforward state-space where the states are the different ways to place $k$ agents into
$|V|$ vertices, one agent per vertex is used. In the {\em start} and {\em goal} states
agent $a_i$ is located at vertices $s_i$ and $g_i$, respectively. Operators
between states are all non-conflicting combinations of actions (including wait)
that can be taken by the agents.

Branching factor in A*-based algorithms is an important measure. Denote $b(a_i)$ the branching factor of single agent $a_i$. Then the {\em effective branching factor} for $k$ agents, denoted by $b$, is $b=\prod_{i=1}^k b(a_i)$. For example, in a 4-connected grid $b(a_i)=5$ for most of agents; an agent can either move in four cardinal directions or wait at its current location. Then $b$, is roughly ${5}^k$; though usually a bit smaller because many possible combinations of moves result in immediate conflicts, especially when the environment is dense with agents.

A simple admissible heuristic that is used within A* for MAPF is to sum the individual heuristics of the single agents such as {\em Manhattan distance} for 4-connected grids or {\em Euclidean distance} for Euclidean graphs~\cite{ryan2008exploiting}. A more-informed heuristic is called the {\em sum of individual costs} heuristic . For each agent $a_i$ we calculate its optimal path cost from its current state (position) $\alpha(a_i)$ to $\alpha_+(a_i)$ assuming that other agents do not exist. Then, we sum these costs over all agents. More-informed heuristics uses forms of pattern-databases~\cite{DBLP:conf/socs/GoldenbergFSHS13,EPEJAIR}.

The most important drawback of A*-based algorithms is they need to tackle with is the branching factor
$b$ of a given state may be exponential in $k$. We briefly summarize attempts to overcome the high branching factor.

{\bf Operator Decomposition (\textsc{Od}).} Instead of moving all the agents to their next positions
at once, agents advance to the next position one
by one in a fixed order within the \textsc{Od} concept. The
original operator for obtaining the next state is thus
decomposed into a sequence of operators for individual agents each of branching factor $b(a_i)$. \textsc{Od} together with a {\em reservation table} enabled computations of next states where agents do not collide with each other in \textsc{Ca*}, \textsc{Hca*}, and \textsc{Whca*} \cite{DBLP:conf/aiide/Silver05}.

Pruning of states by \textsc{Od} with respect to a given admissible heuristic was suggested by Standley in \cite{standley2010finding}.
Two conceptually different states are distinguished - {\em standard} and {\em intermediate}. Intermediate state correspond to the situation when not all the agents finished their move while standard states correspond to states in the original representation with no \textsc{Od}.
The major strength of \textsc{Od} lies in the fact that top-level A* algorithm does not need to distinguish
between standard and intermediate states. The next node for expansion is selected among both standard
and intermediate states while the cost function applies to both types of states. It may thus happen
that a certain intermediate state is not expanded towards a standard state because other states turned out
to be better according to the cost function. Such a kind of search space pruning cannot be done without
operator decomposition as there would be standard states only.

{\bf Independence Detection (\textsc{Id}).} Closely related to \textsc{Od} also introduced in \cite{standley2010finding} is a concept of {\em independence detection} that can also regarded as a branching factor reduction technique. The main idea behind this technique is that difficulty of MAPF solving optimally grows exponentially with the number of agents. It would be ideal, if we could divide the problem into a series of smaller sub problems, solve them independently, and then combine them.

The simple approach, called {\em simple independence detection} (\textsc{Sid}), assigns each agent to a group so that every group consists of exactly one agent. Then, for each of these groups, an optimal solution is found independently. Every pair of these solutions is evaluated and if the two groups’ solutions are in conflict (that is, when a collision of agents belonging to different group occurs), the groups are merged together and a new optimal solution is found for the group (now considering composite search space obtained as a Cartesian product of search spaces of individual groups). If there are no conflicting solutions, the solutions can be merged to a single solution of the original problem. This approach can be further improved by deliberate avoiding of groups merging.

\begin{wrapfigure}{l}{0.5\textwidth}
\begin{center}
\includegraphics[trim={2.5cm 19.3cm 4.5cm 2.8cm},clip,width=1.0\textwidth]{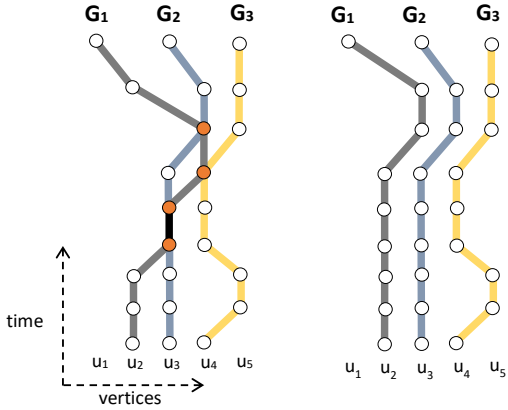}
\end{center}
\caption{Groups $G_1$ conflicts with groups $G_2$ and $G_3$ (left). After replanning $G_1$ independent solutions for $G_1$, $G_2$, and $G_3$ can be merged together. } \label{figure-ID}
\end{wrapfigure}

Generally, each agent has more than one possible optimal path and also group of agents has more that one optimal solution. However, \textsc{Sid} considers only one of these optimal paths/solutions. The improvement of \textsc{Sid} known as {\em independence detection} (\textsc{Id}) is as follows. Let’s have two conflicting groups $G_1$ and $G_2$. First, we try to replan $G_1$ so that the new solution has the same cost but actions that are in conflict with $G_2$ are forbidden. If no such solution is possible, try to similarly replan $G_2$. If this is also not possible, then merge $G_1$ and $G_2$ into a new group. In case either of the replanning was successful, that group needs to be evaluated with every other group again. This can lead to infinite cycle. Therefore, if two groups were already in conflict before, merge them without trying to re-plan. See Figure \ref{figure-ID} for illustration.

Standley uses \textsc{Id} in combination with the A* algorithm. In case A* has a choice between several nodes with the same minimal cost, the one with least amount of conflicts is expanded first. This technique yields an optimal solution that has a minimal number of conflicts with other groups. This property is useful when replanning of a group’s solution is needed.
Both \textsc{Sid} and \textsc{Id} do not solve MAPF on their own, they only divide the problem into smaller sub-problems that are solved by any possible MAPF algorithms. Thus, \textsc{Id} and \textsc{Sid} are general frameworks which can be executed on top of any MAPF solver.

{\bf More A*-based Algorithms.} {\em Enhanced Partial Expansion} (\textsc{Epea*}) \cite{EPEJAIR} avoids
the generation of {\em surplus nodes} (i.e. nodes $n$ with $f(n)>C*$ where $C*$ is the optimal cost; we assume standard A* notation with $f(n)=g(n)+h(n)$) by using {\em a priori} domain knowledge. When expanding a node $n$ \textsc{Epea*} generates only the children $n_c$ with $f(n_c)=f(n)$ and the smallest $f$-value among those children with $f(n_c)>f(n)$ ($\xi$ stands for $f$ in the context of MAPF). The other children of $n$ are discarded. This is done with the help of a domain-dependent {\em operator
selection function} (OSF). The OSF returns the exact list of
operators which will generate nodes $n$ with the desired $f(n)$. Node $n$
is then re-inserted into OPEN setting $f(n)$ to the $f$-value of the next best child of $n$. In this way, EPEA* avoids the generation of surplus nodes and dramatically reduces the number of generated nodes. An OSF for MAPF can be efficiently built as the effect on the $f$-value of moving a single agent in a given direction can be easily computed. For more details see~\cite{EPEJAIR}.

M*~\cite{DBLP:conf/iros/WagnerC11,DBLP:journals/ai/WagnerC15} and its enhanced recursive variant (\textsc{rM*}) are important A*-based algorithms related to \textsc{Id}. M* dynamically changes the {\em dimensionality} and branching factor based on conflicts. The dimensionality is the number of agents that are not allowed to conflict. When a node is expanded, M* initially generates only one child in which each agent takes (one of) its
individual optimal move towards the goal (dimensionality 1). This continues until a conflict occurs between $q \geq 2$ agents at node $n$. At this point, the dimensionality of all the nodes on the branch leading from the root to $n$
is increased to $q$ and all these nodes are placed back in OPEN list. When one of
these nodes is re-expanded, it generates $b^q$ children where the $q$
conflicting agents make all possible moves and the $k-q$ non-conflicting agents
make their individual optimal move. An enhanced variant of M* called \textsc{ODrM*}~\cite{DBLP:conf/icra/FernerWC13} builds
\textsc{rM*} on top of Standley's \textsc{Od} rather than plain A*.

{\bf Increasing Cost Tree Search.} The {\em Increasing Cost Tree Search} algorithm (\textsc{Icts})~\cite{ICTS,DBLP:journals/ai/SharonSGF13} is a two-level MAPF solver which is conceptually different from A*. This algorithm is particularly important as its concepts will be be migrated into the SAT framework we are about to introduce. \textsc{Icts} works as follows.

\begin{wrapfigure}{r}{0.55\textwidth}
\begin{center}
\includegraphics[trim={0cm 0cm 0cm 0.15cm},clip,width=0.5\textwidth]{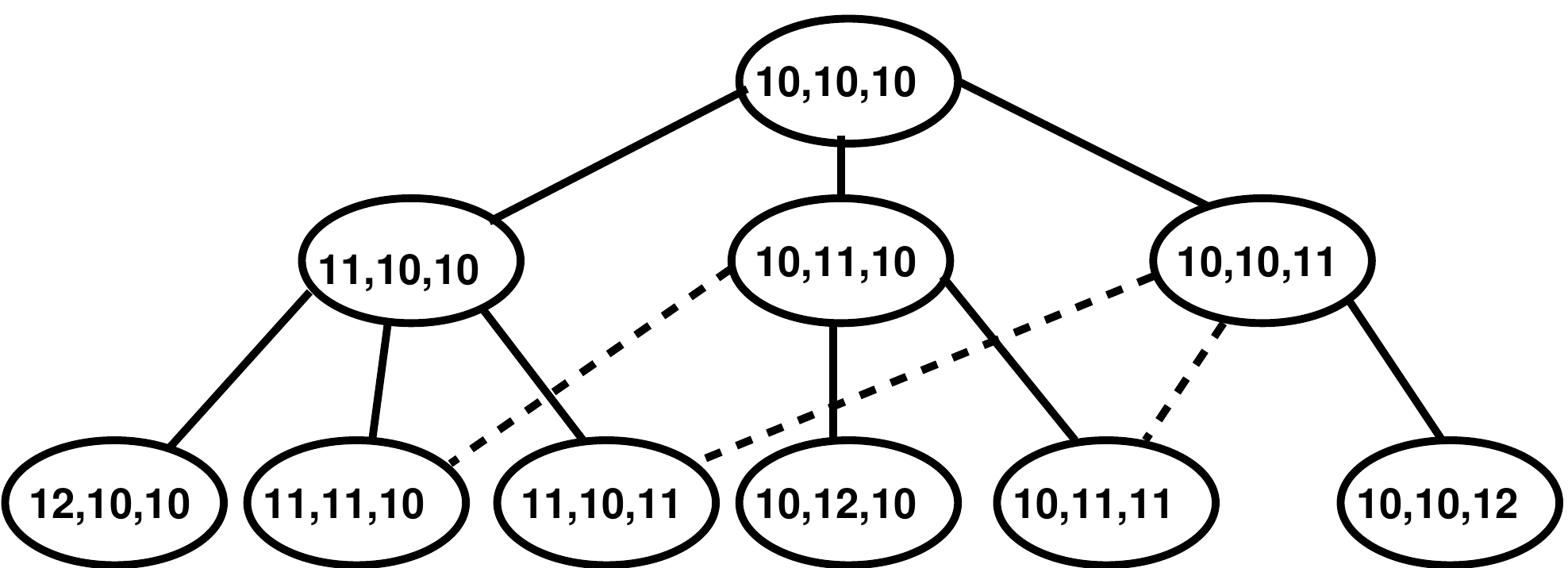}
\end{center}
\caption{Increasing cost tree (\textsc{Ict}) for three agents.} \label{figure-ICT}
\end{wrapfigure}

At its {\em high level}, \textsc{Icts} searches the {\em increasing
cost tree} (\textsc{Ict)}. Every node in the \textsc{Ict} consists of a $k$-ary vector $[\xi(a_1),\xi(a_2),
\ldots, \xi(a_k)]$ which represents {\em all} possible solutions in which the
individual path cost of agent $a_i$ is exactly $\xi(a_i)$. The root of the \textsc{Ict} is
$[\xi^*(a_1),\xi^*(a_2), \ldots,\xi^*(a_k)]$, where $\xi^*(a_i)$ is the optimal individual path cost for
agent $a_i$ ignoring other agents, i.e., it is the length of the shortest path
from $\alpha_0(a_i)$ to $\alpha_+(a_i)$ in $G$.
A child in the \textsc{Ict} is generated by increasing the
cost for one of the agents by 1. An \textsc{Ict} node $[\xi(a_1),\xi(a_2), \ldots \xi(a_k)]$ is a {\em goal} if
there is a complete non-conflicting solution such that the cost of the
individual path for any agent $a_i$ is exactly $\xi(a_i)$. Figure~\ref{figure-ICT}
illustrates an \textsc{Ict} with 3 agents, all with optimal individual path costs of 10.
Dashed lines mark duplicate children which can be pruned. The total cost of a
node is $\sum_{i=0}^k{\xi(a_i)}$. For the root this is exactly $h_{SIC}(\alpha_0)=\sum_{i=0}^k{\xi^*(a_i)}$. Since all nodes at the same height have the same total cost, a breadth-first search of
the \textsc{Ict} will find the optimal solution.

The low level acts as a goal test for the high level. For each
\textsc{Ict} node  $[\xi(a_1),\xi(a_2), \ldots, \xi(a_k)]$  visited by the high level, the low level is invoked.
Its task  is to find a non-conflicting complete solution such that the cost of
the individual path of agent $a_i$ is exactly $\xi(a_i)$. For each agent $a_i$, \textsc{Icts}
stores {\em all} single-agent paths of cost $\xi(a_i)$ in a special compact
data-structure called a {\em multi-value decision diagram}
(MDD)~\cite{DBLP:conf/iccad/SrinivasanKMB90} - MDD will be defined precisely later.

The low level searches the cross product of the MDDs in order to find a set of
$k$ non-conflicting paths for the different agents. If such a non-conflicting
set of paths exists, the low level returns {\em true} and the search halts.
Otherwise, {\em false} is returned and the high level continues to the next
high-level node (of a different cost combination).

\textsc{Icts} also implements various pruning rules to enhance the search. A full study of these pruning rules and
their connection to CSP is provided in~\cite{DBLP:journals/ai/SharonSGF13}.

{\bf Conflict-based Search (\textsc{Cbs}).} Another optimal MAPF solver not based on A* is {\em Conflict-Based
Search} (\textsc{Cbs})~\cite{conf/aaai/CBS12,CBSJUR}. In \textsc{Cbs}, agents are associated with constraints.
A {\em constraint} for agent $a_i$ is a tuple $\langle a_i,v,t \rangle$ where agent $a_i$ is prohibited from
occupying vertex $v$ at time step $t$. A {\em consistent path} for agent $a_i$ is a path that satisfies {\em all} of
$a_i$'s constraints, and a {\em consistent solution} is a solution composed of only consistent paths. 
Once a consistent path has been found for each agent, these paths are {\em validated} with respect to the other agents
by simulating the movement of the agents along their planned paths.

If all agents reach their goal without any conflict the solution is returned. If, however, while performing the
validation, a conflict is found for two (or more) agents, the validation halts and conflict is resolved by adding constraints.
If a \emph{conflict}, $\langle a_i,a_j,v,t \rangle$ is encountered we know that in any valid solution at most one of the
conflicting agents, $a_i$ or $a_j$, may occupy vertex $v$ at time $t$.
Therefore, at least one of the constraints, $\langle a_i,v,t \rangle$ or $\langle a_j,v,t \rangle$, must be satisfied. Consequently, \textsc{Cbs} {\em splits} search into two branches where one of these constraints is valid in each branch.


\subsection{Previous Makespan Optimal Algorithms}

Major development in the makespan optimal MAPF solving has been done over the Boolean satisfiability (SAT) \cite{Biere:2009:HSV:1550723} compilation paradigm. Early works that
compile MAPF to SAT focused on solution improvements in terms of shortening the makespan towards the optimum in {\em anytime} manner \cite{DBLP:conf/pricai/Surynek12}. First, a makespan suboptimal solution of the input MAPF is generated by a fast polynomial rule-based algorithm like \textsc{Bibox} \cite{DBLP:journals/ci/Surynek14} or \textsc{Push-and-Swap} \cite{DBLP:conf/aaai/LunaB11,DBLP:conf/atal/WildeMW13}. Then continuous sub-sequences of time steps in the current solution are replaced by makespan optimal ones. The length of replaced sub-sequences is increased in each iteration of the algorithm until it eventually covers the entire makespan. This ensures that given enough time the algorithm returns makespan optimal solution. A sub-optimal solution is available at any stage of the algorithm.

{\bf \textsc{Inverse} SAT encoding.} Historically the first encoding of MAPF to SAT \textsc{Inverse} relies on {\em log-space} encoded variables \cite{DBLP:books/sp/Petke15} that represent what agent is located in vertex $v$ at each time step $t$ - that is, the inverse $\alpha_t^{-1}:V \rightarrow A \cup \{\bot\}$ of $\alpha$ ($\bot$ stands for empty vertex) is represented using log-space encoded bit-vectors $\mathcal{A}_t^{v} \in \{0,1,2,...,k\}$. MAPF movement rules and state transitions are encoded by a number of constraints over $\mathcal{A}_t^{v}$ - for details see \cite{DBLP:conf/pricai/Surynek12}. Altogether Boolean formula $\mathcal{F}_{\mu}$ is constructed on top $\mathcal{A}_t^{v}$ variables such that it is satisfiable if and only if a solution to the input MAPF of makespan $\mu$ exists. The advantage of this encoding is that the {\em frame problem} \cite{McCHay69} of propagation of agents' positions to the next time step can be easily done by enforcing equalities between $\mathcal{A}_t^{v}$ and $\mathcal{A}_{t+1}^{v}$ (bit-wise equality for all bits of a pair of log-space encoded variables).

Further works in SAT-based approach to MAPF \cite{DBLP:conf/iros/Surynek13,DBLP:conf/sara/Surynek13} omitted the phase in which suboptimal solution was improved and a makespan optimal solution was generated directly instead. The process of finding makespan optimal solution follows the scheme described in Algorithm \ref{MAPF-SAT-MAKE}. Assuming a solvable MAPF a makespan optimal solution is obtained by answering satisfiability of $\mathcal{F}_{\mu_0},\mathcal{F}_{\mu_0+1}, ...$ until a satisfiable formula encountered. The search starts with $\mu_0$, the lower bound on makespan obtained as the length of longest path over all shortest paths connecting starting position $\alpha_0(a_i)$ and goal $\alpha_+(a_i)$ of each agent $a_i$. The first satisfiable $\mathcal{F}_{\mu}$ represents the optimal makespan and an optimal solution can be extracted from its satisfying valuation.

\begin{algorithm}[h]
\begin{footnotesize}
\SetKwBlock{NRICL}{Solve-MAPF-SAT$_{MAKESPAN}$($G=(V,E),A,\alpha_0,\alpha_+)$}{end} \NRICL{
    $paths \gets$ $\{$shortest path from $\alpha_0(a_i)$ to $\alpha_+(a_i)\;|\;i = 1,2,...,k\}$ \\
    $\mu \gets \max_{i=1}^k{(length(paths(a_i)))}$ \\
    \While {$True$}{
        $\mathcal{F}(\mu) \gets$ encode$(\mu,G,A,\alpha_0, \alpha_+)$\\
        $assignment \gets$ consult-SAT-Solver$(\mathcal{F}(\mu))$\\
        \If {$assignment \neq$ UNSAT}{
        	$paths \gets$ extract-Solution$(assignment)$\\
        	\Return $paths$\\
        }
        $\mu \gets \mu + 1$\\
    }
} \caption{Framework of makespan optimal SAT-based MAPF solving} \label{MAPF-SAT-MAKE}
\end{footnotesize}
\end{algorithm}

{\bf \textsc{All-Different} SAT encoding.} The \textsc{All-Different} encoding \cite{DBLP:conf/ictai/Surynek12} again employs log-space representation of variables but position of agent $a_i$ at time step $t$, that is, $\alpha_t$ is represented instead of representing vertex occupancy - that is, variables $\mathcal{D}_t^{a_i} \in V$ are represented using log-space encoding. To ensure that conflicts among agents in vertices do not occur, the \textsc{All-Different} constraint \cite{DBLP:conf/aaai/Regin94} is intorduced for $\mathcal{D}_t^{a_i}$ variables over all agents for each timestep $t$. The advantage of the \textsc{All-Different} encoding is that various efficient encodings of the \textsc{All-Different} \cite{DBLP:conf/fmcad/BiereB08,DBLP:conf/ecai/Surynek12} constraint over bit vectors can be integrated.

{\bf \textsc{Matching} SAT encoding.} The next development has been done in SAT encoding called \textsc{Matching} that separates conflict rules in MAPF and agents transitions between time steps \cite{DBLP:conf/ictai/Surynek14}. Conflict rules are expressed over anonymized agents that are encoded by {\em direct} variables $\mathcal{M}_t^v \in \{True, False\}$.

\begin{wrapfigure}{r}{0.55\textwidth}
\begin{center}
\includegraphics[trim={2.5cm 21cm 5.5cm 3cm},clip,width=0.8\textwidth]{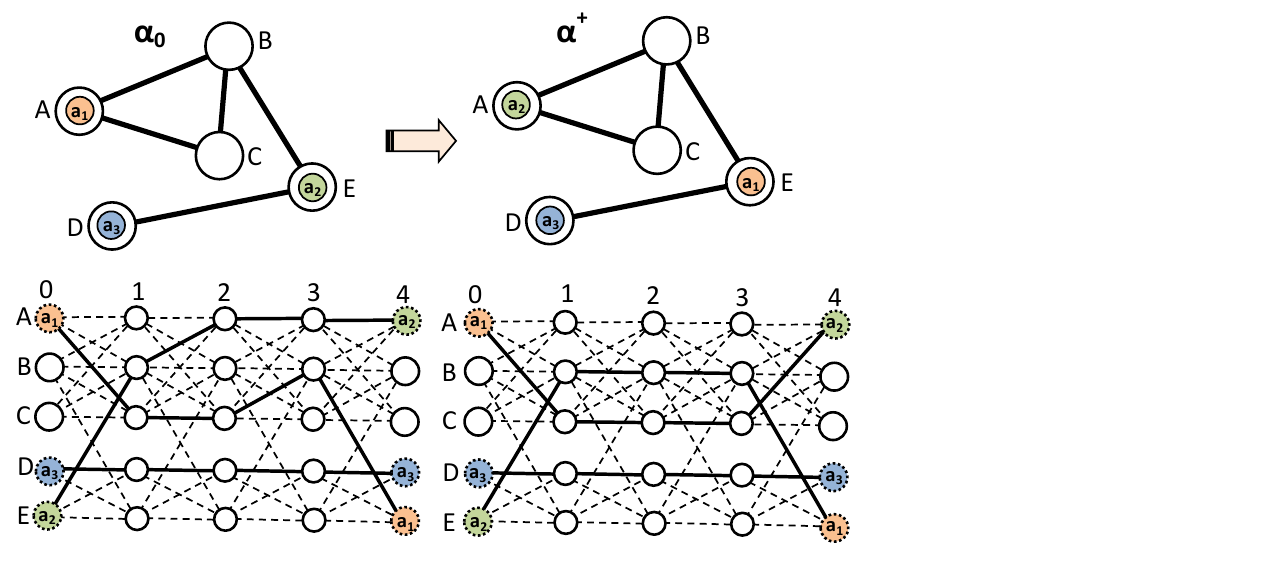}
\end{center}
\caption{Searching of non-conflicting paths over anonymized agents - conflicts are reflected but an agent may end up in the wrong goal (lower right part).} \label{figure-matching}
\end{wrapfigure}

The presence of some agent in vertex $v$ at timestep $t$ is indicated by a single propositional variable ($\mathcal{M}_t^v = True$ if and only if $\exists a_i \in A$ such that $\alpha(a_i)=v$). Using anonymized agents is however not enough as agents may end up in other agent's goal - see Figure \ref{figure-matching}. For transitions where individual agents need to be distinguished, log-space encoded variables $\mathcal{A}_t^v \in \{0,1,2,...,k\}$ represent what agent occupies a given vertex ($\mathcal{A}_t^v$ if and only if $\alpha(a_i)=v$). The advantage of \textsc{Matching} over previous encodings \textsc{Inverse} and \textsc{All-Different} is that movement conflict rules can expressed in a simpler way over direct variables $\mathcal{M}_t^v$ for anonymized agents. Compared to doing so over log-space encoded variables $\mathcal{A}_t^{v}$ or $\mathcal{D}_t^{a_i}$  that distinguish individual agents, smaller formula can be obtained with conflict reasoning over $\mathcal{M}_t^v$.

{\bf \textsc{Direct} SAT encoding.} Lessons taken from the previous development was that introduction of directly encoded variables leads to significant performance improvements although encoding set of states by direct variable is not as space efficient as the in the log-space case. The next encoding purely based on direct variables - called \textsc{Direct} MAPF encoding \cite{DBLP:conf/micai/Surynek14} - introduces a single propositional variable for every triple of agent, vertex, and time step; formally there was a propositional variable $\mathcal{X}(a_i)_t^v$ such that it is $True$ if and only if agent $a_i$ occurs in $v$ at timestep $t$  (some triples may be forbidden as unreachable). In this work we are partly inspired by the \textsc{Direct} encoding as for the of direct variables. 

{\bf ASP, CSP, and ILP approach.} Although lot of work in makespan optimal solving has been done for SAT other compilation-based approaches to MAPF like ASP-based \cite{erdem2013general} and CSP-based \cite{DBLP:conf/icra/Ryan10} exist. Both ASP and CSP offer rich formalism to express various objective functions in MAPF. The ASP-based approach adopts a more specific definition of MAPF where bounds on lengths of paths for individual agents are specified as a part of the input. Except the bound on sum-of-costs the ASP formulation works with other constraints such as {\em no-cycle} (if the agent shall not visit the same part of the environment multiple times), {\em no-intersection} (if only one agent visits each part of the environment), or {\em no-waiting} (when minimization of idle time is desirable). The ASP program for a given variant of MAPF consisting of a combination of various constraints is solver by the \textsc{Clasp} ASP solver \cite{DBLP:conf/lpnmr/GebserKNS07a}.

Ryan in the CSP-based approach focuses on the structure of the underlying graph $G$. The graph is partitioned into {\em halls} (singly-linked chain of vertices with any number of entrances and exits) and {\em cliques} (represents large open spaces with many entrances and exists) commonly refered to as sub-graphs. The plan is searched using CSP techniques over an abstract graph whose nodes are represented by sub-graphs. Specific properties of different sub-graphs are reflected in constraints - for example, agents in a clique sub-graph never exceed the capacity and the agents preserve their ordering in a hall sub-graphs. The resulting CSP is eventually solved using the \textsc{Gecode} solver \cite{Tack:PhD:2009}.

The similarity of MAPF and multi-commodity flows is studied in \cite{DBLP:conf/aaai/YuL13} where each agent is regarded as a different commodity. Depths of the multi-commodity flow are associated with individual time steps of MAPF solution. Finding optimal solutions of MAPF with respect to various objective functions can be then modeled as finding optimal solution of Integer Linear Programming (ILP) problem \cite{DBLP:conf/icra/YuL13}.

\section{The New SAT-based Solvers}

SAT solvers \cite{Biere:2009:HSV:1550723} encompass Boolean variables and answer binary questions. The challenge is to apply SAT for MAPF where there is a cumulative cost function. This challenge is stronger for the sum-of-costs variant where each agent has its own cost. We first recall main ideas of SAT encodings for makespan. Then, we present our SAT encoding for sum-of-costs.

\subsection{SAT Encoding for Optimal Makespan}

\begin{wrapfigure}{r}{0.5\textwidth}
\begin{center}
\includegraphics[trim={4.5cm 20.5cm 10.5cm 4.5cm},clip,width=0.4\textwidth]{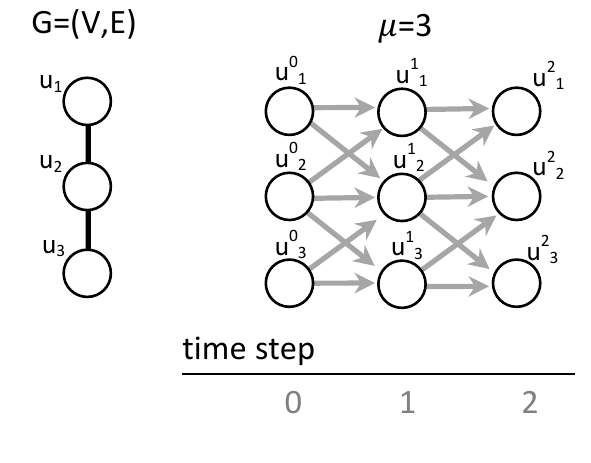}
\end{center}
\caption{An example of time expansion graph: input graph (left) and its expansion for 3 steps.} \label{figure-TEG}
\end{wrapfigure}

A {\em time expansion graph} (denoted TEG) is a basic concept used in SAT solvers for makespan optimal MAPF solving \cite{DBLP:conf/ictai/Surynek14}. We use it too in the sum-of-costs variant below. A TEG is a directed acyclic graph (DAG). First, the set of vertices of the underlaying graph $G$ is duplicated for all time-steps from 0 up to the given makespan bound $\mu$. Then, possible actions (move along edges or wait) are represented as directed edges between successive time steps. Figure~\ref{figure-TEG} shows a graph and its TEG for time steps 0, 1 and 2 (vertical layouts).

It is important to note that in this example (1) horizontal edges in TEG correspond to {\em wait} actions. (2) diagonal moves in TEG correspond to real moves. Formally a TEG is defined as follows:

\begin{definition}
\textit{Time expansion graph} (TEG) of depth $\mu$ is a digraph $(V',E')$ derived from G=(V,E) where $V'=\{u^t_j\;|\;t=0,1,...,\mu \wedge u_j \in V\}$ and $E' = \{{(u_j^t,u_k^{t+1})}\;|\;t=0,1,...,\mu-1\ \wedge (\{u_j,u_k\} \in E \vee j=k) \}$. \label{def:modifiedExpansionGraph}
\end{definition}

The encoding for MAPF introduces TEGs for indivudual agents. That is, we have $TEG_i=(V_i,E_i)$ for each agent $i \in \{1,2,...,k\}$. Directed non-conflicting paths in TEGs correspond to valid non-conflicting movements of agents in the underlying graph G. The existence of non-conflicting paths in TEGs will be encoded as satisfiability of a Boolean formula. We will describe in more details encoding style used in the \textsc{Direct} encoding.

Boolean variables and constraints (clauses) for a single time-step $t \in \{0,1,...,\mu\}$ in $TEG_i$ in order to represent any possible location of agent $a_i$ at time $t$; that is, we have $\mathcal{X}(a_i)_v^t$. Boolean variables for all TEGs together represent all possible arrangements of agents from timestep $0$ up to timestep $\mu$. It is ensured by constraints that arrangements of agents in consecutive time steps of TEGs correspond to valid actions: agent $a_i$ can appear in vertex $u_j^t$ if it can move there from the previous time step in $TEG_i$ along a directed edge, that is, if $a_i$ is in some $u_k^{t-1}$ such that $(u_k^{t-1},u_j^t) \in E_i$. We also have inter-TEG constrains ensuging that agents do not collide with each other (detailed list of constraints will be introduced for the sum-of-costs variant).

Given a desired makespan $\mu$, formula $\mathcal{F}_{\mu}$ represents the question of whether there is a collection of non-conflicting directed paths in $TEG_1$,...,$TEG_k$ of depth $\mu$ such that the first arrangement equals to $\alpha_0$ and the last one equals $\alpha_+$. The search for optimal makespan is done by iteratively incrementing $\mu=0,1,2...$ until a satisfiable formula $\mathcal{F}_{\mu}$ is obtained as shown in Algorithm \ref{MAPF-SAT-MAKE}.

This process ensures finding makespan optimal solution in case of a solvable input MAPF instance since satisfiability of $\mathcal{F}_{\mu}$ is a non-decreasing function of $\mu$. It is important to note that solvability of a given MAPF can be checked in advance by a fast polynomial algorithm like \textsc{Push-and-Rotate} \cite{DBLP:conf/atal/WildeMW13}. More information on SAT encoding for the makespan variant can be found, e.g. in \cite{DBLP:conf/ictai/Surynek14,DBLP:conf/pricai/Surynek14,DBLP:conf/micai/Surynek14}. The detailed transformation of a question of whether there are non-conflicting paths in TEGs will shown in following sections.

\section{Basic-SAT for Optimal Sum-of-costs}

The general scheme described above for finding optimal makespan is to convert the optimization problem (finding minimal makespan) to a sequence of decision problems (is there a solution of a given makespan $\mu$). The decision problem was: is there a solution of makespan $\mu$,  and the sequence of decision problems was to increment $\mu$ until the minimal makespan is found (this works due to monotinicity of existence of solution w.r.t. increasing makespan; wait action can prolong a solution arbitrarily). The questions are {\em which decision problem to encode}, {\em how to encode it},  and {\em how to devise an appropriate sequence of these decision problems} that  will guarantee a solution to the the optimization problem at hand.

In the makespan MAPF variant, the {\em numeric objective function} to minimize, i.e., the makespan $\mu$  corresponds directly to the number of time expansions of the underlaying graph $G$ in TEGs. Thus, the decision problem was: is there a solution in a TEGs of depth $\mu$. This decision problem can be regarded as a question: are there non-conflicting directed paths in TEGs that interconnects agents' starting positions and goals. The existence of such paths is then encoded into a Boolean formula.

We apply the same scheme for finding optimal sum-of-costs, converting it to a sequence of decision problems -- is there a solution of a given sum-of-costs $\xi$ where the decision problem is whether there is a solution of sum-of-costs $\xi$, and the sequence of decision problems is to increment $\xi$ until finding the minimal sum-of-costs is found. Again the solvability of a MAPF instance is monotonic w.r.t. increasing sum-of-costs hence the above incemental strategy works.

It is important to note that incremental strategy to obtain the optimal value of the objective function is suitable only when the cost of query is exponential in $\mu$ or $\xi$ (in case of uniform query cost different strategies like binary search would be more suitable). This roughly holds in the MAPF as increasing $\mu$ corresponds to adding a fresh time step in TEGs which is reflected in the encoded Boolean formula by adding a number of variables and constraints proportional to the size of $G$. Since the runtime of a SAT solver is exponential in the size of the input formula in the worst case we have that runtime for answering $\mathcal{F}_{\mu}$ is exponential in $\mu$ in the worst case. In such a setup with incremental strategy, the cost/runtime of the last query is roughly the same as the total cost/runtime of previous queries. As we will see later, the same applies also for the sum-of-costs $\xi$.

However, encoding the decision problem for the sum-of-costs is more challenging than the makespan case, because one needs to both bound the sum-of-costs, but also to predict how many time expansions are needed. We address this challenge by using two key
novel techniques described next: {\bf (1)} Cardinality constraint for bounding $\xi$ and {\bf (2)} Bounding the Makespan.

\begin{itemize}
\item{
\noindent {\bf Cardinality constraints.} This is a technique from the SAT literature that enables counting and bounding a numeric cost in a Boolean formulate \cite{DBLP:conf/cp/BailleuxB03,DBLP:conf/cp/SilvaL07,DBLP:conf/cp/Sinz05}. This enables encoding a constraint that bounds the sum of cost inside Boolean formulae. Typically the encoding of cardinality constraints is based on simulation of arithmetic circuits for calculating summations. While most of logic circuits assume binary encoding of input values where weights of individual bits/propositional variables is determined by their position, here we work with unary encoding of inputs where each single propositional variable contributes by 1 to the overall sum (see Section \ref{sec:detailedDescription} for details).
}

\item{
\noindent {\bf Upper bound on the required time expansions.} We show below how to compute for a given sum of cost value $\xi$ a value $\mu$ such that all possible solutions with sum-of-costs $\xi$ must be possible for a makespan of at most $\mu$ (details in Section~\ref{sec:boundingTheMakespan}). This enables encoding the decision problem of whether there is a solution of sum-of-costs $\xi$ by using a SAT encoding similar to the makespan encoding with $\mu$ time expansions. In other words, it will be sufficient to use TEGs of depth
$\mu$ in order to represent all solutions that fits under the given sum-of-costs $\xi$.
}
\end{itemize}

Next, we explain each of these techniques in detail, along with theoretical analysis and additional implementation details.

\subsection{Cardinality Constraint for Bounding $\xi$}

The SAT literature offers a technique for encoding a \textit{cardinality constraint}~\cite{DBLP:conf/cp/SilvaL07,DBLP:conf/cp/Sinz05},
which allows calculating and bounding a numeric cost within the Boolean formula. Formally, for a bound $\lambda \in \mathbb{N}$ and a set of Boolean variables $X=\{x_1,x_2,...,x_k\}$ the \textit{cardinality constraint} $\leq_{\lambda}{\{x_1,x_2,...,x_k\}}$ is satisfied iff the number of variables from the set $X$ that are set to \texttt{TRUE} is $ \leq \lambda$. There are various ways how to encode cardinality constraints in Boolean formulae. The standard approach is to simulate arithmetic circuits \cite{DBLP:conf/cp/BailleuxB03} inside the formula. 
Arithmetic circuits for cardinality constraints usually assume unary encoding of inputs where each propositional variable (bit) from $X$ contributes by 1 to the sum.

In our SAT encoding, we bound the sum-of-costs by mapping every agent's action to a Boolean variable, and then encoding a cardinality constraint on these variables. Thus, one can use the general structure of the makespan SAT encoding (which iterates over possible makespans), and add such a cardinality constraint on top. Next we address the challenge of how to connect these two factors together.

\subsection{Bounding the Makespan for the Sum of Costs}
\label{sec:boundingTheMakespan}

Next, we compute how many time expansions $\mu$ are needed to guarantee that if a solution with sum-of-costs $\xi$ exists then it will be found within at most $\mu$ time expansions. In other words, in our encoding, the values we give to $\xi$ and $\mu$ must fulfill the following requirement:

\vspace{0.25cm}
\noindent{\em {\bf R1:} All possible solutions with sum-of-costs $\xi$ must be possible for a makespan of at most $\mu$.}
\vspace{0.25cm}

To find a $\mu$ value that meets R1 for given $\xi$, we require the following definitions. Let $\xi_0(a_i)$ be the cost of the shortest individual path for agent $a_i$ ($\xi_0(a_i) = length(path(a_i))$), and let $\xi_0=\sum_{a_i\in A} \xi_0(a_i)$. $\xi_0$ was called the {\em sum of individual costs} (SIC)\cite{DBLP:journals/ai/SharonSGF13}. $\xi_0$ is an admissible heuristic for optimal sum-of-costs search algorithms, since $\xi_0$ is a lower bound on the minimal sum-of-costs. $\xi_0$ is calculated by relaxing the problem by omitting the other agents (collisions with them).  Similarly, we define $\mu_0=\max_{a_i\in A} \xi_0(a_i)$.  $\mu_0$ is length of the {\em longest} of the shortest individual paths and is thus a lower bound on the minimal makespan. Finally, let $\Delta$ be the extra cost over SIC (as done in \cite{DBLP:journals/ai/SharonSGF13}). That is, let $\Delta = \xi - \xi_0$.

\newtheorem{proposition}{Proposition}
\begin{proposition}
For makespan $\mu$ of any solution with sum-of-costs $\xi$, R1 holds for $\mu \leq \mu_0 + \Delta$. \label{prop:upperbound}
\end{proposition}

\noindent {\bf Proof:} The worst-case scenario, in terms of makespan, is that all the $\Delta$ extra moves belong to a single agent. Given this scenario, in the worst case, $\Delta$ is assigned to the agent with the largest shortest-path. Thus, the resulting path of that agent would be $\mu_0+\Delta$, as required. $\Box$
\vspace{0.25cm}

Using Proposition~\ref{prop:upperbound}, we can safely encode the decision problem of whether there is a solution with sum-of-costs $\xi$ by using $\mu=\mu_0+\Delta$ time expansions, knowing that if a solution of cost $\xi$ exists then it will be found within $\mu=\mu_0+\Delta$ time expansions. In other words,  Proposition~\ref{prop:upperbound} shows relation of both parameters $\mu$ and $\xi$ which will be both changed by changing $\Delta$. Algorithm~\ref{MAPF-SAT-SOC} summarizes our optimal sum-of-costs algorithm.

In every iteration, $\mu$ is set to $\mu_0+\Delta$ (Line 4) and the relevant TEGs of depth $\mu$ (described below) for the various agents are built. Using TEGs of individual agents a formula  $\mathcal{F}(\mu,\Delta)$ is constructed that encodes a decision problem whether there is a solution with sum-of-costs $\xi$ and makespan $\mu$. Afterwards the formula is queried to the SAT solver (Line 8).
The first iteration starts with $\Delta=0$. If such a solution exists, it is returned. Otherwise $\xi$ is incremented by one, $\Delta$ and consequently $\mu$ are modified accordingly and another iteration of SAT consulting is activated.

\begin{algorithm}[t!]
\begin{footnotesize}
\SetKwBlock{NRICL}{Solve-MAPF-SAT$_{SUM-OF-COSTS}$(MAPF $\Sigma=(G=(V,E),A,\alpha_0,\alpha_+)$)}{end} \NRICL{
	\If {$\Sigma$ is unsolvable}{
		\Return (Solution not found)\\
   } \Else { 
   $\mu_0=\max_{a_i\in A} \xi_0(a_i)$\\
	$\Delta \gets 0$\\
    \While {$True$} {
        $\mu \gets \mu_0 + \Delta$\\
         \For {each agent $a_i$}{
            $TEG_i(\mu) \gets$ Construct-TEG($\mu, \xi_0(a_i), \alpha_0(a_i), \alpha_+(a_i), G$)\\
            }
            $\mathcal{F}(\mu,\Delta) \gets$ encode($\mu, \Delta, \alpha_0,\alpha_+, TEG_1(\mu),...,TEG_k(\mu)$) \\
        $assignment \gets$ Consult-SAT-Solver($\mathcal{F}(\mu,\Delta)$)\\
        \If {$assignment \neq UNSAT$}{
            $paths \gets$ extract-Solution($assignment$)\\
            \Return $paths$\\
        }
        \Else {
            $\Delta \gets \Delta + 1$\\        
        }
    }
    }
} \caption{Basic SAT-based sum-of-costs optimal MAPF solving.}\label{MAPF-SAT-SOC}
\end{footnotesize}
\end{algorithm}

\begin{proposition}
The algorithm SAT consult is sound and complete. \label{prop:algorithm}
\end{proposition}

\noindent {\bf Proof:} This algorithm clearly terminates; for unsolvable instances after the initial solvability test; for solvable MAPF instances as we start seeking a solution of $\xi=\xi_0$ ($\Delta=0$) and increment $\Delta$ (which increments $\xi$ and $\mu$ as well) to all possible values. Hence we eventually encounter $\Delta$ ($\xi$ and $\mu$) for which $\Sigma$ is solvable and valid solution is calculated and returned. $\Box$
\vspace{0.25cm}

The initial unsolvability check of an MAPF instance can be done by any polynomial-time complete sub-optimal algorithm such as \textsc{Push-and-Rotate}~\cite{DBLP:journals/jair/WildeMW14}.

\subsection{Efficient Use of the Cardinality Constraint}

The complexity of encoding a cardinality constraint depends linearly in the
number of constrained variables~\cite{DBLP:conf/cp/SilvaL07,DBLP:conf/cp/Sinz05}.
Since each agent $a_i$ must move at least $\xi_0(a_i)$, we can reduce the
number of variables counted by the cardinality constraint by only counting the
variables corresponding to extra movements over the first $\xi_0(a_i)$ movement
$a_i$ makes. We implement this by introducing a TEG for a given agent $a_i$
(labeled $TEG_{i}$).

\begin{figure}[t]
\centering
\includegraphics[trim={4.0cm 20.9cm 4.0cm 4.5cm},clip,width=1.0\textwidth]{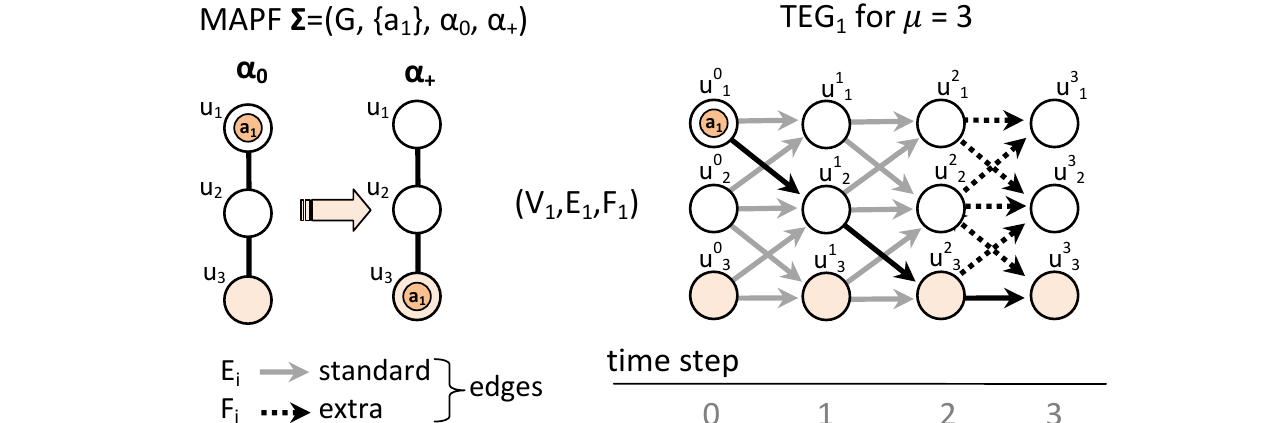}
\caption{A TEG for an agent that needs to go from $u_1$ to $u_3$.}
\label{figure-TEGi}
\end{figure}

$TEG_i$ differs from TEG (Definition~\ref{def:modifiedExpansionGraph}) in that
it distinguishes between two types of edges: $E_i$ and $F_i$. $E_i$ are
(directed) edges whose destination is at time step $\leq{}\xi_0(a_i)$. These
are called {\em standard edges}. $F_i$ denoted as \textit{extra edges} are
directed edges whose destination is at time step $>{}\xi_0(a_i)$. A formal construction
of the time expansion graph is shown using pseudo-code as Algorithm \ref{TEG-EXPAND}.

\begin{algorithm}[t!]
\begin{footnotesize}
\SetKwBlock{NRICL}{Construct-TEG($\mu$, $\xi_0$, $s$, $g$, $(G=(V,E)$)}{end} \NRICL{
    $V_i \gets \emptyset$ \\
    $E_i \gets \emptyset$ \\
    $F_i \gets \emptyset$ \\
    \For {$u_i \in V$}{
        $E \gets E \cup \{u_i,u_i\}$ /* adding loops to ensure the frame axiom */ \\
    }
    \For {$t \in \{0,1,...,\mu\}$} {  
    	  $V_i \gets V_i \cup \{u_j^t\;|\;u_j \in V\}$\\
    }
    \For {$t \in \{0,1,...,\mu-1\}$} {      
    	  \For {each $\{u_j,u_l\} \in E$}{  	      
    	      \If {$t \leq \xi_0$}{
  	          $E_i \gets E_i \cup \{u_j^t,u_l^{t+1}\}$    	      
    	      }
    	      \Else {
    	          \If {$\{u_j,u_l\} \neq \{t\}$}{
    	              $F_i \gets F_i \cup \{u_j^t,u_l^{t+1}\}$
    	          }
    	          \Else {
    	              $E_i \gets E_i \cup \{u_j^t,u_l^{t+1}\}$
    	          }
    	      }
    	  }
   }
   \Return ($TEG_i = (V_i,E_i,F_i)$)\\
} \caption{Construction of the time expansion graph.}\label{TEG-EXPAND}
\end{footnotesize}
\end{algorithm}

Figure~\ref{figure-TEGi} shows an underlying graph for agent $a_1$ (left) and
the corresponding $TEG_1$. Note that the optimal solution of cost 2 is denoted
by the diagonal path of the TEG. Edges that belong to $F_i$ are those that
their destination is time step 3 (dotted lines), only these edges can contribute to the
sum-of-costs above $\xi_0(a_1)=2$. That is, we will only bound the number of extra
edges (they sum up to $\Delta$) making the encoding of the cardinality constraint
more efficient.

\subsection{Detailed Description of the SAT Encoding}
\label{sec:detailedDescription}

Agent $a_i$ must go from its initial position to its goal within $TEG_i$. This
simulates its location in time in the underlying graph $G$. That is, the task
is to find a path from $\alpha_0^0(a_i)$ to $\alpha_+^\mu(a_i)$ in $TEG_i$. The
search for such a path will be encoded within the Boolean formula. Additional
constraints will be added to capture all movement constraints such as
\textit{collision avoidance} etc. And, of course, we will encode the
cardinality constraint that the number of extra edges must be exactly $\Delta$.

We want to ask whether a sum-of-costs solution of $\xi$ exist. For this we
build $TEG_i$ for each agent $a_i \in A$ of depth $\mu_0 + \Delta$.
We use $V_i$ to denote the set of vertices in $TEG_i$ that agent $a_i$ might
occupy during the time steps. Next we introduce the basic Boolean encoding (denoted
\texttt{BASIC-SAT}) which has the following Boolean variables.
\vspace{0.25cm}

\noindent {\bf 1:)}  $\mathcal{X}^t_j(a_i)$ for every $t \in \{0,1,...,\mu \}$
and $u^t_j \in V_i$ -- Boolean variable of whether agent $a_i$ is in vertex
$u_j$ at time step $t$.
\vspace{0.25cm}

\noindent {\bf 2:)} $\mathcal{E}^t_{j,l}(a_i)$ for every $t \in \{0,1,...,\mu-1
\}$ and $(u^t_j,u^{t+1}_l) \in (E_i \cup F_i)$
-- Boolean variables that model transition of agent $a_i$ from vertex $u_j$ to
vertex $u_k$ through any edge (standard or extra) between time steps $t$ and
$t+1$ respectively.
\vspace{0.25cm}

\noindent {\bf 3:)}  $\mathcal{C}^t(a_i)$ for every $t \in \{0,1,...,\mu-1 \}$
such that there exist $u^t_j \in V_i$ and $u^{t+1}_l \in V_i$ with
$(u^t_j,u^{t+1}_l) \in F_i$ -- Boolean variables that model cost of movements
along {\bf extra edges} (from $F_i$) between time steps $t$ and $t + 1$.
\vspace{0.25cm}

We now introduce constraints on these variables to restrict illegal values as
defined by our variant of MAPF. Other variants may use a slightly different
encoding but the principle is the same. Let $T_\mu = \{0, 1, ..., \mu-1\}$.
Several groups of constraints are introduced for each agent $a_i \in A$ as
follows:
\vspace{0.25cm}

\noindent{\bf C1:} If an agent appears in a vertex at a given time step, then
it must follow through exactly one adjacent edge into the next time step. This
is encoded by the following pseudo-Boolean constraint \cite{Biere:2009:HSV:1550723},
which is posted for every $t \in T_\mu$ and $u_j^t \in V_i$:

\begin{equation}
   {  \mathcal{X}^t_j(a_i) \Rightarrow \sum_{(u^t_j,u^{t+1}_l) \in E_i \cup F_i }{\mathcal{E}^t_{j,l}(a_i)=1}
   }
\end{equation}

The above pseudo-Boolean can be translated to clauses in multiple ways. One simple and efficient way is to rewrite the constraint as follows:

\begin{equation}
   {  \mathcal{X}^t_j(a_i) \Rightarrow \bigvee_{(u^t_j,u^{t+1}_l) \in E_i \cup F_i }{\mathcal{E}^t_{j,l}(a_i),}
      }\label{eq:basic-start}
\end{equation}
\begin{equation}
   {  \bigwedge_{(u^t_j,u^{t+1}_l),(u^t_j,u^{t+1}_h) \in E_i \cup F_i \wedge l<h}{\neg\mathcal{E}^t_{j,l}(a_i) \vee \neg\mathcal{E}^t_{j,h}(a_i)}
      }\label{eq:basic-1}
\end{equation}
\vspace{0.25cm}

\noindent{\bf C2:} Whenever an agent occupies an edge it must also enter it
before and leave it at the next time-step. This is ensured by the following
constraint introduced for every $t \in T_\mu$ and $(u^t_j,u^{t+1}_l) \in E_i
\cup F_i$:
\begin{equation}
   {  \mathcal{E}^t_{j,l}(a_i) \Rightarrow \mathcal{X}^t_j(a_i) \wedge \mathcal{X}^{t+1}_l(a_i)
    } \label{eq:basic-2}
\end{equation}
\vspace{0.25cm}

\noindent{\bf C3:} The target vertex of any movement except wait action must be
empty. This is ensured by the following constraint introduced for every $t \in
T_\mu$ and $(u^t_j,u^{t+1}_l) \in E_i \cup F_i$ such that $j\neq l$.
\begin{equation}
   {  \mathcal{E}^t_{j,l}(a_i) \Rightarrow \bigwedge_{a_h \in A \wedge a_h \neq a_i \wedge u^t_j \in V_h}{\neg \mathcal{X}^t_j(a_h)}
    } \label{eq:basic-3}
\end{equation}
\vspace{0.25cm}

\noindent{\bf C4:} No two agents can appear in the same vertex at the same time
step. Again this can be expressed by the following pseudo-Boolean constraint
for every vertex $u_j \in V$ and $t \in T_\mu$:

\begin{equation}
   {  \sum_{a_i \in A \wedge u^t_j \in V_i}{\mathcal{X}^t_j(a_i)} \leq 1
    }
\end{equation}

Equivalently this can be expressed by following binary clauses  for every pair of of agents $a_i,a_l \in A$ such that $i \neq h$:
\begin{equation}
   {  \bigwedge_{u^t_j \in V_i\cap V_h}{\neg \mathcal{X}^t_j(a_i) \vee \neg \mathcal{X}^t_j(a_h)}
    }\label{eq:basic-4}
\end{equation}
\vspace{0.25cm}

\noindent{\bf C5:} Whenever an extra edge is traversed the cost needs to be
accumulated. In fact, this is the only cost that we accumulate as discussed
above. This is done by the following constraint for every $t \in T_\mu$ and
extra edge $(u^t_j,u^{t+1}_l) \in F_i$:
\begin{equation}
   { \mathcal{E}^t_{j,l}(a_i) \Rightarrow \mathcal{C}^t(a_i)
    }\label{eq:basic-5}
\end{equation}
\vspace{0.25cm}

\noindent{\bf C6:} The cost of wait action followed by a non-wait action need to be accumulated. This is ensured by the following constraint for every $t \in T_\mu$:

\begin{equation}
   { \mathcal{C}^t(a_i) \Rightarrow \bigwedge_{\sigma \in \{0,1,...,t-1\} \wedge \{(u^\sigma_j,u^{\sigma+1}_l) \in F_i\} \neq \emptyset }\mathcal{C}^\sigma(a_i)
    }\label{eq:basic-6}
\end{equation}

\noindent{\bf C7: Cardinality constraint.} Finally the bound on the total cost
needs to be introduced. Reaching the sum-of-costs of $\xi$ corresponds to
traversing exactly $\Delta$ extra edges from $F_i$. The following cardinality
constrains ensures this:
\begin{equation}
   { \leq_{\Delta}\Big\{
   \begin{tabular}{c}
   $\mathcal{C}^t(a_i)\;|\;i=1,2,...,n \wedge t \in T_{\mu} \wedge \{(u^t_j,u^{t+1}_l) \in F_i\} \neq \emptyset$
   \end{tabular}
   \Big\}
    }
   \label{eq:basic-end}
\end{equation}
\vspace{0.25cm}

\noindent{\bf Final formula.}  The resulting Boolean formula that is a
conjunction of $C1 \ldots C7$ will be denoted as $\mathcal{F}_{BASIC}(\mu, \Delta)$ and is the one
that is consulted by Algorithm~\ref{MAPF-SAT-SOC} (lines 11-12).
\vspace{0.25cm}

The following proposition summarizes the correctness of our encoding.
\begin{proposition}
MAPF $\Sigma=({G=(V,E)},A,\alpha_0,\alpha_+)$ has a \textit{sum-of-costs}
solution of $\xi$ if and only if $\mathcal{F}_{BASIC}(\mu, \Delta)$ is
satisfiable. Moreover, a solution of MAPF $\Sigma$ with the sum-of-costs of
$\xi$ can be extracted from the satisfying valuation of
$\mathcal{F}_{BASIC}(\mu, \Delta)$ by reading its
$\mathcal{X}^t_j(a_i)$ variables.
\end{proposition}

\noindent {\bf Proof:} The direct consequence of the above definitions is that
a valid solution of a given MAPF $\Sigma$ corresponds to non-conflicting paths
in the TEGs of the individual agents. These non-conflicting paths further
correspond to the satisfying variable assignment of
$\mathcal{F}_{BASIC}(\mu, \Delta)$, i.e., that there are $\Delta$ extra
edges in TEGs of depth $\mu=\mu_0+\Delta$. $\Box$

\begin{proposition}
Let $D$ be the maximal degree of any vertex in $G=(V,E)$ and let $k$ be the number
of agents. If $|E|\geq \mu$ and $k \geq D$ then the number of clauses
in $\mathcal{F}_{BASIC}(\mu, \Delta)$ is $O(\cdot k^2\cdot \mu \cdot |E|)$,
and the number of variables is $O(k \cdot \mu \cdot |E|)$.
 \end{proposition}
 
\noindent {\bf Proof:} The components of  $\mathcal{F}_{BASIC}(\mu,
\Delta)$  is described in equations~\ref{eq:basic-start}-- \ref{eq:basic-end}.
Equations~\ref{eq:basic-start} and \ref{eq:basic-1} introduce at most $O(k\cdot \mu \cdot |V| \cdot D^2)$
clauses. Equation~\ref{eq:basic-2} introduces at most $O(k \cdot \mu \cdot |E|)$ clauses. Equation~\ref{eq:basic-3} introduces at most $O(k^2 \cdot \mu \cdot |E|)$ clauses. Equation~\ref{eq:basic-4} introduces at most $O(k^2 \cdot \mu \cdot |V|)$ clauses. Equation~\ref{eq:basic-5} introduces at most $O(k \cdot \mu \cdot |E|)$  clauses. Equation~\ref{eq:basic-6} introduces at most $O(k \cdot \mu^2)$ clauses. And finally equation~\ref{eq:basic-end} introduces at most $O(k\cdot \mu \cdot \Delta)$ clauses, since a cardinality constraint checking that $n$ variables has a cardinality constraint of $k$ requires $O(n\cdot m)$ clauses~\cite{DBLP:conf/cp/Sinz05}.

Summing all the above results in a total of $O(k \cdot \mu \cdot (|V| \cdot D^2 + k \cdot |E| + \mu + \Delta))$. If we assume that
$k \geq D$, $|E| \geq \mu$, and that $k \cdot|E| \geq \mu$ (by definition $\mu \geq \Delta$) then the number of clauses is $O(k^2
\cdot \mu \cdot |E|)$. The number of variables is easily computed in a similar
way. $\Box$


\vspace{0.25cm}
Various optimizations of the encoding could be done at the level of using alternative encodings of the cardinality constraints and/or by eliminating edge variables $\mathcal{E}^t_{j,l}(a_i)$ via equivalence $\mathcal{E}^t_{j,l}(a_i) \Leftrightarrow \mathcal{X}^t_{j}(a_i) \wedge \mathcal{X}^{t+1}_{l}(a_i)$. We observed that these optimizations represent minor changes in the overall size and efficiency of the encoding.

\section{Experimental Evaluation}
\noindent

\noindent
We experimented on 4-connected grids with randomly placed obstacles
~\cite{DBLP:conf/aiide/Silver05,standley2010finding} and on \textit{Dragon Age}
maps~\cite{CBSJUR,sturtevant2012benchmarks}. Both settings are a standard MAPF
benchmarks. The initial position of the agents was randomly
selected. To ensure solvability the goal positions were selected by performing
a long \textit{random walk} from the initial arrangement.

\begin{figure*}[t]
\centering
\includegraphics[trim={2cm 8cm 16cm 2.5cm},clip,width=0.80\textwidth]{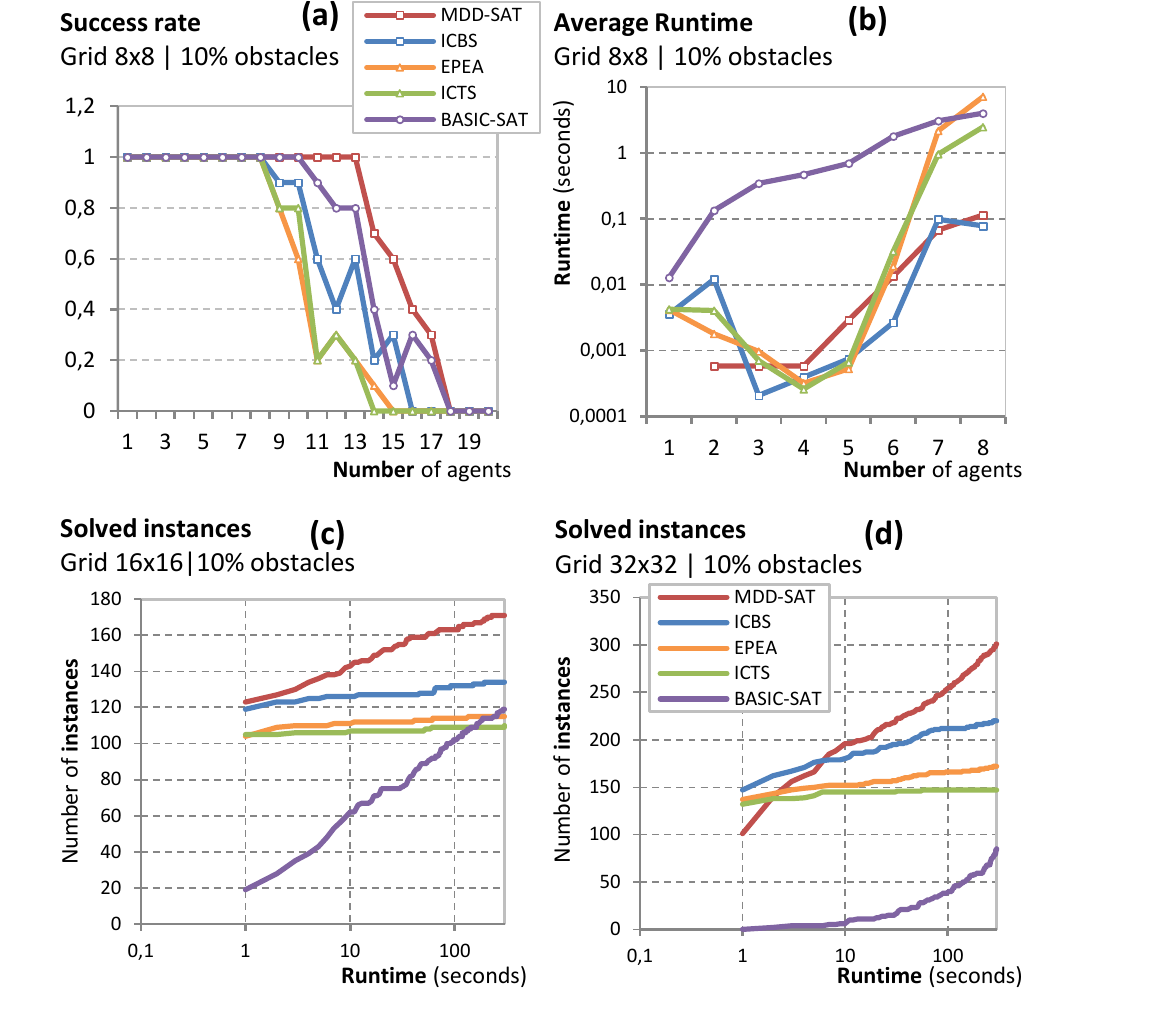}
\caption{Results on $8{}\times{}8$ grid (left). Number of solved instances in
the given runtime on $16{}\times 16$ and $32{}\times{}32$ grids.
(right)}\label{figure-grids}
\end{figure*}


We compared our SAT solvers to several state-of-the-art search-based
algorithms: the \textit{increasing cost tree search} - ICTS
\cite{DBLP:journals/ai/SharonSGF13}, {\em Enhanced Partial Expansion A*} -
EPEA*~\cite{EPEJAIR} and \textit{improved conflict-based search} -
ICBS \cite{DBLP:conf/ijcai/BoyarskiFSSTBS15}.  For all the search
algorithms we used the best known setup of their parameters and enhancements
suitable for solving the given instances over 4-connected grids.


The SAT approaches were implemented in C++. The implementation consists of a
top level algorithm for finding the optimal sum-of-costs $\xi$ and {\em CNF}
formula generator \cite{Biere:2009:HSV:1550723} that prepares input formula for
a SAT solver into a file. The SAT solver is an external module our this
architecture. We used \texttt{Glucose
3.0}~\cite{DBLP:conf/ijcai/AudemardS09,DBLP:conf/sat/AudemardLS13} which is a
top performing SAT solver in the \textit{SAT Competition}
\cite{DBLP:journals/aim/JarvisaloBRS12,DBLP:conf/ictai/Surynek14}.

The cardinality constraint was encoded using a simple standard circuit based
encoding called \textit{sequential counter}~\cite{DBLP:conf/cp/Sinz05}. In our
initial testing we considered various encodings of the cardinality constrain
such as those discussed in
\cite{DBLP:conf/cp/BailleuxB03,DBLP:conf/cp/SilvaL07}. However, it turned out
that changing the encoding has a minor effect.\footnote{Due to the knowledge of
lower bounds on the sum-of-costs, the number of variables involved in the
cardinality constraint is relatively small and hence the different encoding
style has not enough room to show its benefit.}

ICTS and ICBS were implemented in C\#, based on their original implementation
(here we used a slight modification in which the target vertex of a move must
be empty). All experiments were performed  on a Xeon 2Ghz, and on Phenom II
3.6Ghz, both with 12 Gb of memory.

\subsection{Square Grid Experiments}

We first experimented on  $8{}\times{}8$, $16{}\times{}16$, and
$32{}\times{}32$ grids with 10\% obstacles while varying the number of agents
from 1 up to the number where at least one solver was able to solve an instance
(in case of the $8{}\times{}8$ grid this is 20 agents; and 32 and 58 in case of
$16{}\times{}16$ and $32{}\times{}32$ grids respectively). For each number of
agents 10 random instances were generated.

Figure~\ref{figure-grids} presents results where each algorithm was given a
time limit of 300 seconds (as was done
by~\cite{DBLP:journals/ai/SharonSGF13,DBLP:conf/ijcai/BoyarskiFSSTBS15,SharonSFS15}).
The leftmost plot (Plot (a)) shows the {\em success rate} (=percentage out of
given 10 random instances solved within the time limit) as a function of the
number of agents for the $8{}\times{}8$ grid (higher curves are better). The
next plot (Plot (b)) reports the average runtime for instances that were solved
by all algorithms (lower curves are better). Here, we required $100\%$ success
rate for all the tested algorithms to be able to calculate average runtime;
this is also the reason why the number of agents is smaller. The two right
plots visualize the results on $16{}\times{}16$ grid (Plot (c)) and
$32{}\times{}32$ grid (Plot (d)) but in a different way. Here, we present the
number of instances (out of all instances for all number of agents) that each
method solved ($y$-axis) as a function of the elapsed time ($x$-axis). Thus, for
example Plot (c) says that MDD-SAT was able to solve 145 instances in time less
than 10 seconds (higher curves are better).

The first clear trend is that MDD-SAT significantly outperforms BASIC-SAT in
all aspects. This shows the importance of developing enhanced SAT encodings for
the MAPF problem. The performance of the BASIC-SAT encoding compared to the
search-based algorithm degrades as the size of the grids grow larger: in the
8x8 grids it is second only to MDD-SAT, in the 16x16 grid it is comparable to
most search-based algorithms, and in the 32x32 grid it is even substantially
worse. For the rest of the experiments we did not activate BASIC-SAT.

In addition, a prominent trend observed in all the plots is that MDD-SAT has
higher success rate and solves more instances than all other algorithms. In
particular, in on highly constrained instances (containing many agents) the
MDD-SAT solver is the best option.

However, on the $32{}\times{}32$ grid (rightmost figure) for easy instances
when the available runtime was less than 10 seconds, MDD-SAT was weaker than
the search-based algorithms. This is mostly due to the architecture of the
MDD-SAT solver which has an overhead of running the external SAT solver and
passing input in the textual form to it. This effect is also seen in the 8x8
plot (Plot (b)) as these were rather easy instances (solved by all algorithms)
and the extra overhead of activating the external SAT solver did not pay off.

\begin{figure}[t]
\centering
\includegraphics[trim={2cm 22.4cm 8cm 2.5cm},clip,width=0.80\textwidth]{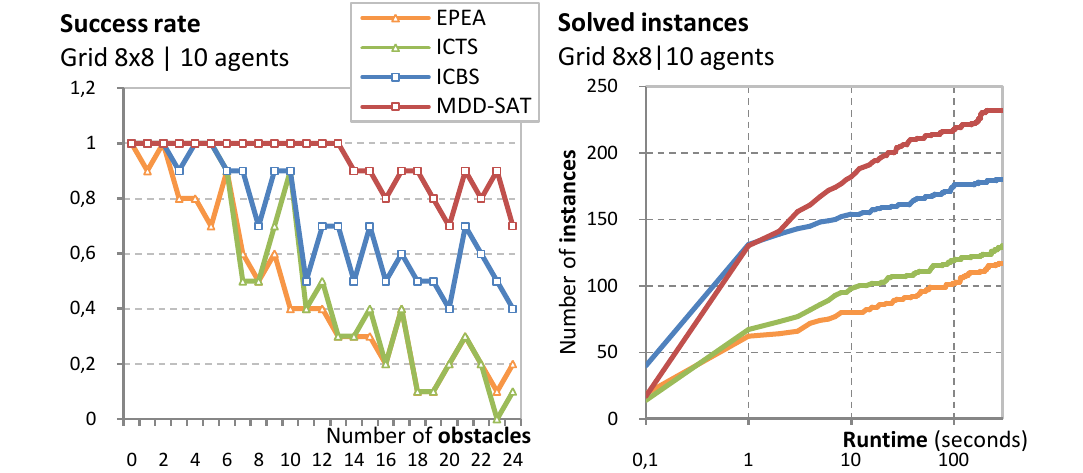}
\caption{Success rate and runtime on the $8 \times 8$ grid with increasing
number of obstacles (out of 64 cells).} \label{figure-obstacles}
\end{figure}

Next, we varied the number of obstacles for the $8{}\times{}8$ grid with 10
agents to see the impact of shrinking free space and increasing the frequency
of interactions among agents. Results are shown in
Figure~\ref{figure-obstacles}. Again, MDD-SAT clearly solves more instances
over all settings. MDD-SAT was always faster except for some easy instances
(that needed up to 1 second) where ICBS was slightly faster which is again due
to the overhead in setup of the SAT solving by an external solver.
Interestingly, increasing the number of obstacles reduces the number of open
cells. This is an advantage for the SAT formula generator in MDD-SAT as the
formula has less variables and constraints. By contrast, the combinatorial
difficulty of the instances increases with adding obstacles for all the solvers
as it means that the graphs gets denser and harder to solve.

\subsection{Results on the Dragon Age Maps}

\begin{figure}[t]
\centering
\includegraphics[trim={1.5cm 22.5cm 7cm 2.7cm},clip,width=0.80\textwidth]{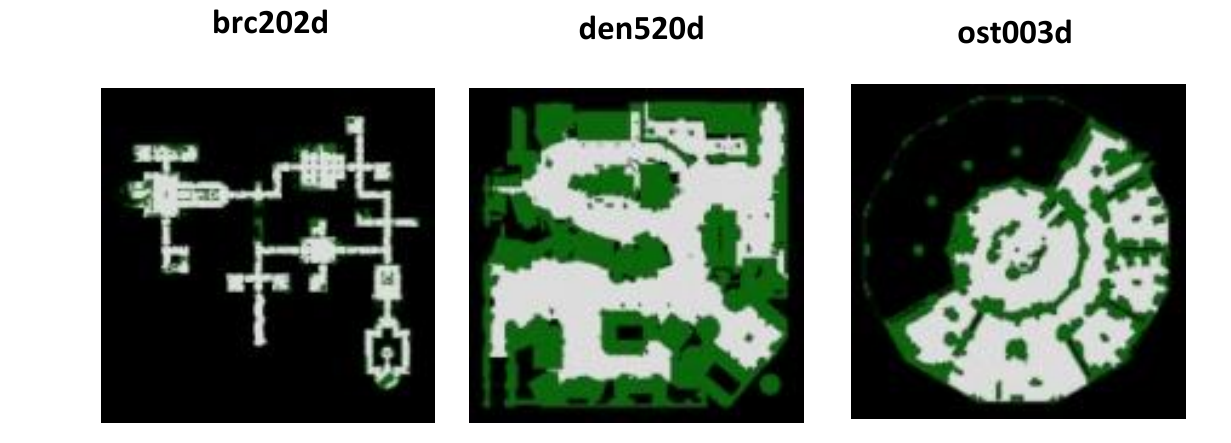}
\caption{Three structurally diverse \texttt{Dragon-Age} maps used in the experimental evaluation. This selection includes: narrow corridors in \texttt{brc202d}, large open space in \texttt{den520d}, and open space with almost isolated rooms in \texttt{ost003d}.} \label{figure-maps}
\end{figure}

Next, we experimented on three structurally different Dragon-Age maps -
\texttt{ost003d}, \texttt{den520d}, and \texttt{brc202d}, that are commonly
used as testbeds \cite{DBLP:journals/ai/SharonSGF13,EPEJAIR,
DBLP:conf/ijcai/BoyarskiFSSTBS15} - see Figure \ref{figure-maps}. On these maps
we only evaluated the  most efficient algorithms, namely, MDD-SAT, ICTS, and
ICBS. Generally, in these maps there is a large number of open cells but the
graph is sparse with agents but there are topological differences.
\texttt{brc202d} has many narrow corridors. \texttt{ost003d} consists of few
open areas interconnected by narrow doors. Finally, \texttt{den520d} has wider
open areas.

To obtain instances of various difficulties we varied the distance between
start and goal locations. Ten random instances were generated for each distance
in the range: $\{8,16,24,\ldots,320\}$ in order to have instances of
different difficulties (total of 400 instances). With larger distances, the
problems are more difficult as the the probability for interactions (avoidance)
among agents increases as they need to travel through a larger part of the
graph.

\begin{figure}[t]
\centering
\includegraphics[trim={2cm 17.2cm 8cm 2.5cm},clip,width=0.80\textwidth]{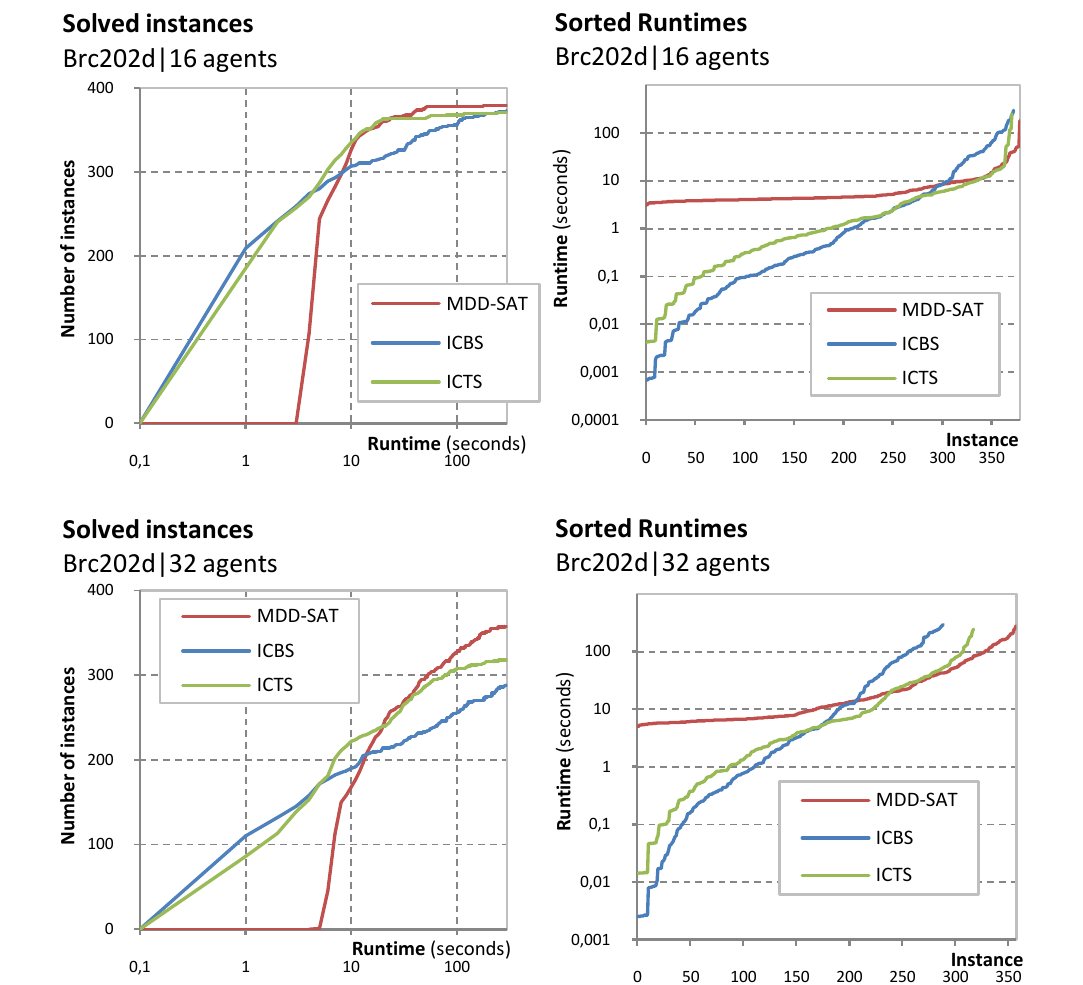}
\caption{Results for dragon age map \texttt{brc202d} with 16 and 32 agents. The left part shows the number of instances ($y$-axis) a solver manages to solve in the given timeout ($x$-axis). The right part shows all the runtimes for a given solver sorted in the ascending order.} \label{figure-map-brc202d}
\end{figure}

\begin{figure}[t]
\centering
\includegraphics[trim={2cm 17.2cm 8cm 2.5cm},clip,width=0.80\textwidth]{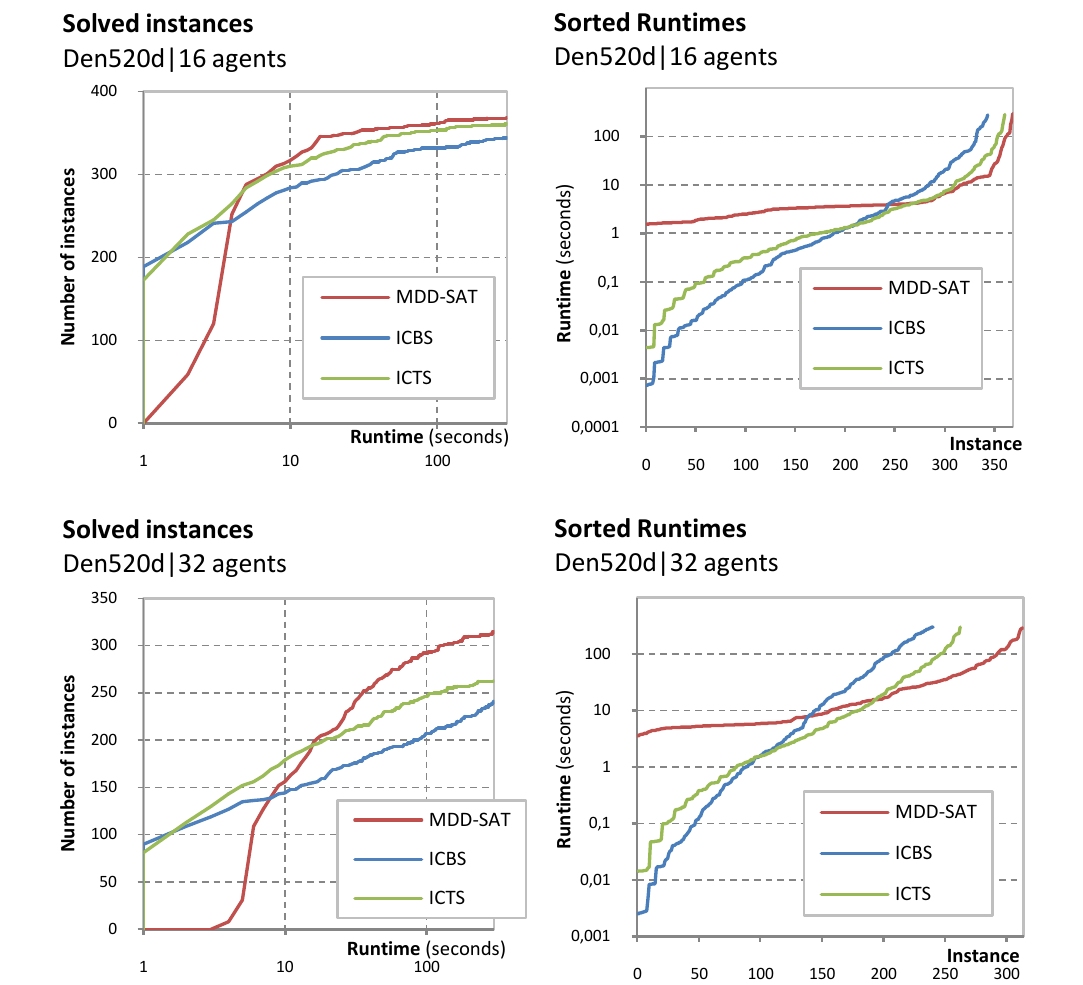}
\caption{Results for dragon age map \texttt{den520d} with 16 and 32 agents. MDD-SAT is the best option on hard instances with more agents.} \label{figure-map-den520d}
\end{figure}

\begin{figure}[t]
\centering
\includegraphics[trim={2cm 17.2cm 8cm 2.5cm},clip,width=0.80\textwidth]{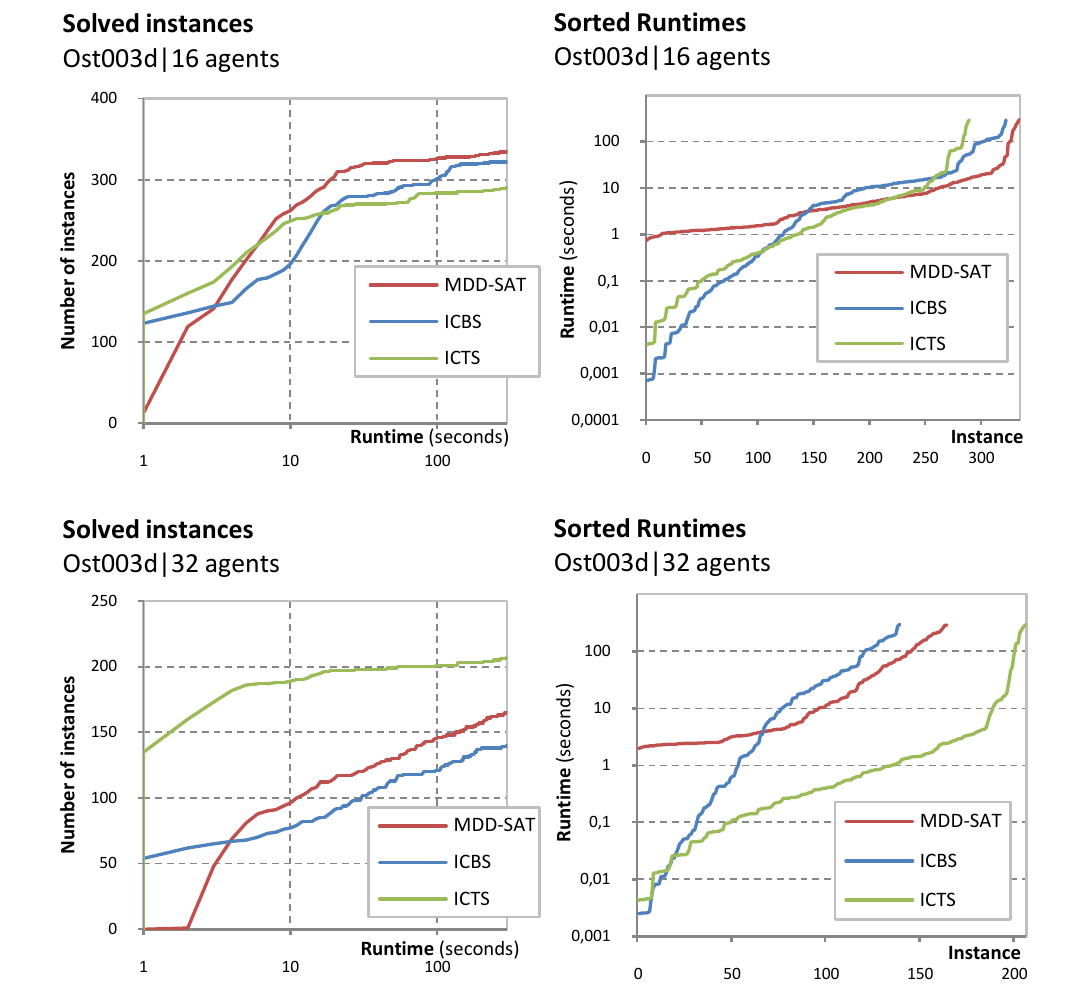}
\caption{Results for dragon age map \texttt{ost003d} with 16 and
32 agents. Although MDD-SAT performs as best with 16 agents, it gets
outperformed in the case with 32 agents by ICTS. This case shows that there is
no universal winner among the tested algorithms.} \label{figure-map-ost003d}
\end{figure}

The results for the three Dragon-Age maps are shown in
Figure~\ref{figure-map-brc202d} (\texttt{brc202d}),
Figure~\ref{figure-map-den520d} (\texttt{den520d}), and
Figure~\ref{figure-map-ost003d} (\texttt{ost003d}). Two setups were used for
each map - one with 16 agents, the other with 32 agents. The left plot of each
figure shows the number of solved instances ($y$-axis) as a function of the
elapsed time ($x$-axis). Again, higher curves correspond to better
performance). The right plot is interpreted as follows. For each solver the 400
instances are ordered in increasing order of their solution time (this has
strong correlation with the distance between the start and goal
configurations). Thus, the numbers in the $x$-axis give the relative location
(out of the 400) in this sorted order. The $y$-axis gives the actual running
time for each instance. Here, lower curves correspond to better performance.

All these figures show a similar clear trend with the exception of
\texttt{ost003d} with 32 agents (discussed below). On the easy instances where
little time is required (left of the figures), MDD-SAT is not the best. But,
for the harder instances that need more time (right of the figures), MDD-SAT
clearly outperform all the other solvers.

Intuitively, one might think that the search-based solvers will have an
advantage in these domains since they contain many open spaces (low
combinatorial difficulty) while the MDD-SAT approach will suffer here as it
will need to generate a large number of formulae (as the domains are large).
This might be true for the easy instances.  Nevertheless, the effectiveness of
MDD-SAT was clearly seen on the harder instances where generating the formulae
and the external time to activate the architecture of the SAT solver seemed to
pay off. This trend was also seen in in the case of small densely occupied
grids discussed above.

The \texttt{ost003d} map with 32 agents is the only case where MDD-SAT was
outperformed by ICTS. This is probably due to the specific structure of
\texttt{ost003d} which has a number of isolated open spaces. This gives an
advantage to ICTS with relatively many agents (32) as conflicts mostly occur at
the exits/doors of the open areas. ICTS handles this on a per-agent cost basis
while the other solvers are less effective here.

The entire set of experiments show a clear trend.  For the easy instances when
a small amount of time is given the search-based algorithm may be faster. But,
given enough time MDD-SAT is the correct choice, even in the large maps where
it has an initial disadvantage. One of the reasons for this is modern SAT
solvers have the ability to learn and improve their speed during the process of
answering a SAT question. But, this learning needs sufficient time and large
search trees to be effective. By contrast, search algorithms do not have this
advantage.


\subsection{Size of the Formulae}

Concrete runtimes for 10 instances of {\em ost003d} are given in
Table~\ref{table_ost003}. MDD-SAT solves the hardest instance (\#1) while other
solvers ran out of time. The right part of the table illustrates the cumulative
size of the formulae generated during the solving process. Although the map is
much larger than the square grids, the size of formulae is comparable to the
densely occupied grid. This is because
$\xi_0$ is a good lower bound of the optimal cost in the sparse maps.

The observation from this experiments is that the large underlaying graph does not
necessarily imply generating of large Boolean formulae in the MDD-SAT solving
process. Though in harder scenarios (where start and goals are far apart) large
formulae are eventually generated but still do not represent any significant
disadvantage for the MDD-SAT solver according to presented measurements. We observed that generating large formulae takes considerable portion of the total runtime (up to $10\%$-$30\%$) within the MDD-SAT solver. Hence efficient implementation of this part of the solver has significant impact on the overall performance.

\begin{table}
\centering
\includegraphics[trim={2cm 21.4cm 9.7cm 1.7cm},clip,width=0.75\textwidth]{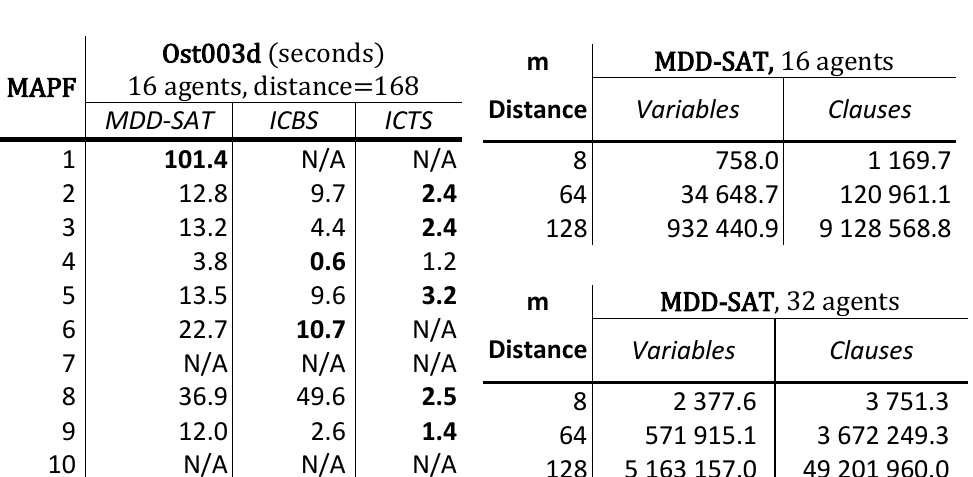}
\caption{Runtime for 10 instances (left) and the average size
of the MDD-SAT formulae for \texttt{ost003d} (right)}\label{table_ost003}
\end{table}

\section{Summary and Conclusion}

We summarized how to migrate techniques from search-based optimal MAPF solvers to
SAT-based method. The outcome is the first SAT-based solver for the sum-of-costs
variant of MAPF. The new solver was experimentally
compared to the state-of-the-art search-based solvers over a variety of domains
- we tested 4-connected grids with random obstacles and large maps from
computer games. We have seen that the SAT-based solver is a better option in hard scenarios
while the search-based solvers may perform better in easier cases.

Nevertheless, as previous authors
mentioned~\cite{CBSJUR,DBLP:conf/ijcai/BoyarskiFSSTBS15} there is no universal
winner and each of the approaches has pros and cons and thus might work best in
different circumstances. For example, ICTS was best on \texttt{ost003d} with 32
agents. This calls for a deeper study of various classes of MAPF instances and
their characteristics and how the different algorithms behave across them. Not
too much is known at present to the MAPF community on these aspects.

There are several factors behind the performance of the SAT-based approach:
clause learning, constraint propagation, good implementation of the SAT solver.
On the other hand, the SAT solver does not understand the structure of the
encoded problem which may downgrade the performance. Hence, we consider that
implementing techniques such as learning directly into the dedicated MAPF
solver may be a future direction. Finally, migrating of other ideas from both
classes of approaches might further improve the performance.

Another interesting future direction is  to consider how additional techniques from search-based MAPF solvers can be used to further improve our SAT-based search, e.g., incorporating our SAT-based solver in the meta-agent conflict-based search framework as a low-level solver~\cite{GUNI2}. This includes techniques for decomposing the problem to subproblems~\cite{standley2010finding} and incorporating our SAT-based solver in the meta-agent conflict-based search framework as a low-level solver~\cite{GUNI2}.


\section{Acknowledgments}

This research has been supported by the Czech Science Foundation (grant application number 19-17966S).

\section{References}

\bibliographystyle{plain}
\bibliography{migrating-MAPF-2SAT_Journal-2018}

\end{document}